\documentclass{fairmeta}

\usepackage{graphicx}    
\usepackage{natbib}      
\usepackage{appendix}    
\usepackage{booktabs}    
\usepackage{url}         
\usepackage{hyperref}    
\usepackage{cleveref}    
\usepackage{wrapfig}     
\usepackage{amssymb}     

\usepackage{xcolor}      
\usepackage{tcolorbox}   
\usepackage{colortbl}    

\usepackage[frozencache,cachedir=minted-cache]{minted} 
\setminted{fontsize=\scriptsize}                      

\usepackage{subfig}        
\usepackage{subcaption}    
\usepackage{caption}       

\usepackage{tabularx}      
\usepackage{longtable}     
\usepackage{makecell}      
\usepackage{arydshln}      
\usepackage{siunitx}       

\usepackage{tikz}
\usepackage{tikzsymbols}

\usepackage{amsthm}        
\usepackage{mathrsfs}      

\usepackage{enumerate}     
\usepackage{enumitem}      

\usepackage{fontawesome}   
\usepackage{pifont}        

\usepackage{ragged2e}      
\usepackage{listings}      
\usepackage[normalem]{ulem} 
\useunder{\uline}{\ul}{}   
\usepackage{fancyvrb}      

\usepackage[noframe]{showframe} 
\usepackage{placeins}      

\usepackage{epigraph}      
\usepackage{bookmark}      

\crefname{section}{section}{sections}        
\Crefname{section}{Section}{Sections}        
\crefname{figure}{figure}{figures}         
\Crefname{figure}{Figure}{Figures}         
\crefname{table}{table}{tables}          
\Crefname{table}{Table}{Tables}          

\definecolor{RoyalBlue}{RGB}{65,105,225}
\definecolor{green(ncs)}{rgb}{0.0, 0.62, 0.42}
\definecolor{peach}{rgb}{1.0, 0.9, 0.71}
\definecolor{palesilver}{rgb}{0.79, 0.75, 0.73}
\definecolor{palegoldenrod}{rgb}{0.93, 0.91, 0.67}
\definecolor{oldlace}{rgb}{0.99, 0.96, 0.9}
\definecolor{navajowhite}{rgb}{1.0, 0.87, 0.68}
\definecolor{mediumchampagne}{rgb}{0.95, 0.9, 0.67}
\definecolor{macaroniandcheese}{rgb}{1.0, 0.74, 0.53}
\definecolor{lemonchiffon}{rgb}{1.0, 0.98, 0.8}
\definecolor{radicalred}{rgb}{1.0, 0.21, 0.37}
\definecolor{pinkpearl}{rgb}{0.91, 0.67, 0.81}
\definecolor{deepsaffron}{rgb}{1.0, 0.6, 0.2}
\definecolor{electricpurple}{rgb}{0.75, 0.0, 1.0}
\definecolor{amber(sae/ece)}{rgb}{1.0, 0.49, 0.0}
\definecolor{amethyst}{rgb}{0.6, 0.4, 0.8}
\definecolor{arylideyellow}{rgb}{0.91, 0.84, 0.42}
\definecolor{awesome}{rgb}{1.0, 0.13, 0.32}
\definecolor{ballblue}{rgb}{0.13, 0.67, 0.8}
\definecolor{calpolypomonagreen}{rgb}{0.12, 0.3, 0.17}
\definecolor{electricindigo}{rgb}{0.44, 0.0, 1.0}
\definecolor{fulvous}{rgb}{0.86, 0.52, 0.0}
\definecolor{mikadoyellow}{rgb}{1.0, 0.77, 0.05}
\definecolor{pumpkin}{rgb}{1.0, 0.46, 0.09}
\definecolor{jsonkey}{RGB}{153,0,0}     
\definecolor{jsonstring}{RGB}{36,36,255} 
\definecolor{jsonnumber}{RGB}{0,128,0}  
\definecolor{jsonnull}{RGB}{128,128,128} 
\definecolor{metablue}{rgb}{0.00, 0.47, 1.00}
\definecolor{mikadoyellow}{rgb}{1.0, 0.66, 0.07}
\definecolor{fluorescentorange}{rgb}{1.0, 0.75, 0.0}
\definecolor{candypink}{rgb}{0.89, 0.44, 0.48}
\definecolor{mediumcarmine}{rgb}{0.69, 0.25, 0.21}
\definecolor{phthaloblue}{rgb}{0.0, 0.06, 0.54}
\definecolor{teal}{rgb}{0.0, 0.5, 0.5}

\definecolor{Green}{RGB}{34,139,34}
\hypersetup{
    colorlinks=true,
    citecolor=RoyalBlue,
    linkcolor=RoyalBlue,
    filecolor=magenta,
    urlcolor=RoyalBlue
}

\newcommand{\dependencyarrow}{\textcolor{orange}{$\rightarrow$}}

\NewDocumentCommand\grayboxicon{}{{\raisebox{-0.9mm}{\includegraphics[scale=0.020]{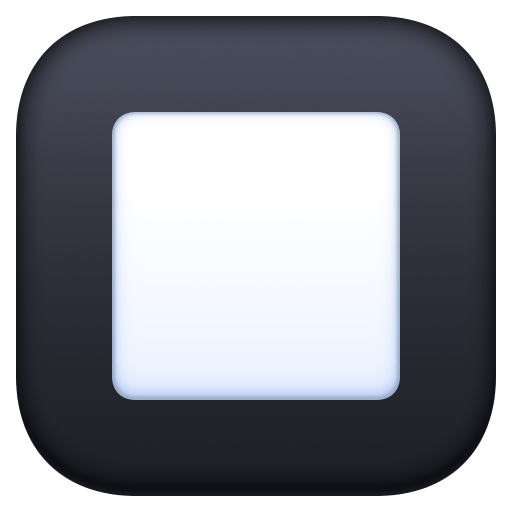}}}}
\NewDocumentCommand\blackboxicon{}{{\raisebox{-0.9mm}{\includegraphics[scale=0.020]{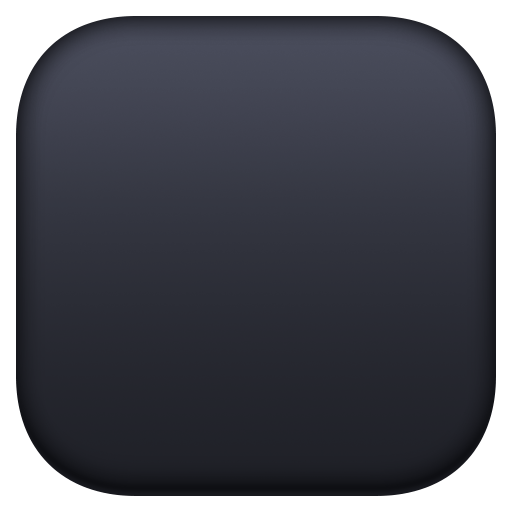}}}}
\NewDocumentCommand\whiteboxicon{}{{\raisebox{-0.9mm}
{\includegraphics[scale=0.020]{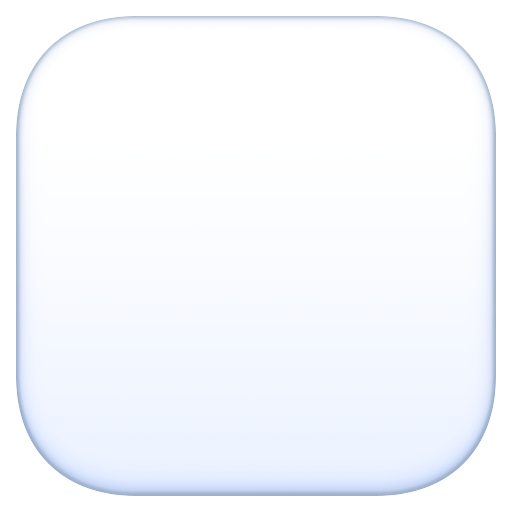}}}}
\NewDocumentCommand\statisticsicon{}{{\raisebox{-0.9mm}{\includegraphics[scale=0.022]{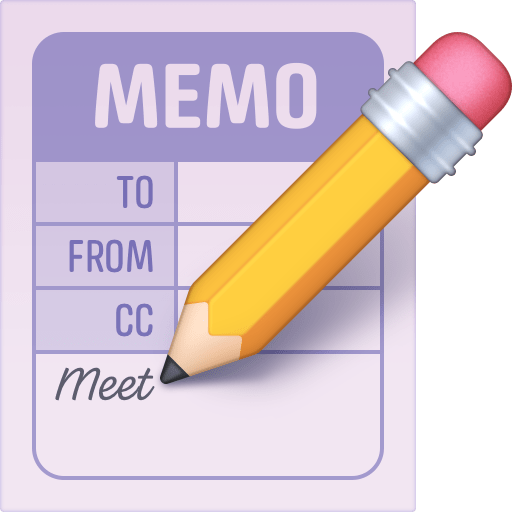}}}}
\NewDocumentCommand\humanicon{}{{\raisebox{-1.1mm}{\includegraphics[scale=0.022]{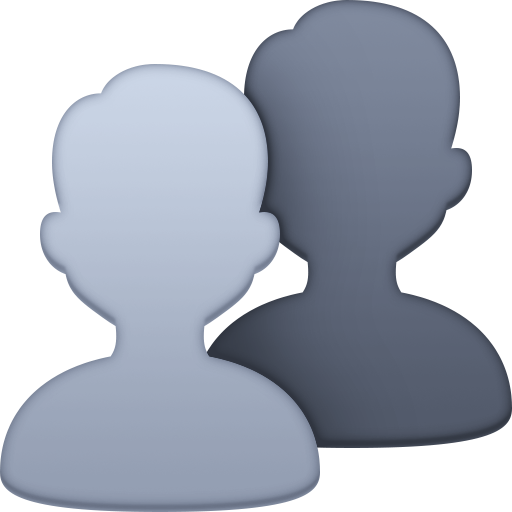}}}}
\NewDocumentCommand\lighticon{}{{\raisebox{-0.9mm}{\includegraphics[scale=0.025]{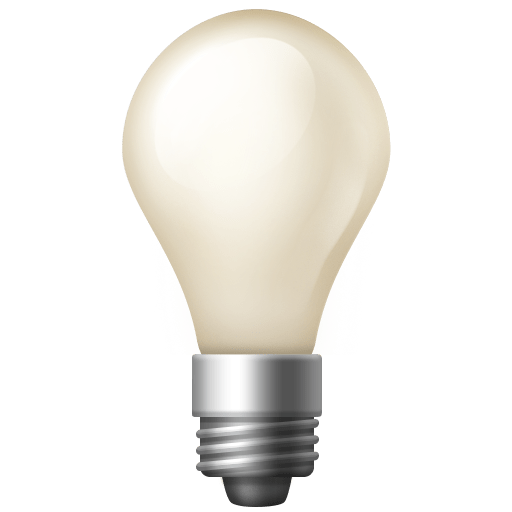}}}}

\lstdefinelanguage{json}{
    basicstyle=\ttfamily\small,
    showstringspaces=false,
    breaklines=true,
    frame=lines,
    backgroundcolor=\color{metablue!10},  
    morestring=[b]",
    literate=
     *{0}{{{\color{jsonnumber}0}}}{1}
      {1}{{{\color{jsonnumber}1}}}{1}
      {2}{{{\color{jsonnumber}2}}}{1}
      {3}{{{\color{jsonnumber}3}}}{1}
      {4}{{{\color{jsonnumber}4}}}{1}
      {5}{{{\color{jsonnumber}5}}}{1}
      {6}{{{\color{jsonnumber}6}}}{1}
      {7}{{{\color{jsonnumber}7}}}{1}
      {8}{{{\color{jsonnumber}8}}}{1}
      {9}{{{\color{jsonnumber}9}}}{1}
      {:}{{{\color{jsonkey}{:}}}}{1}
      {,}{{{\color{jsonkey}{,}}}}{1}
      {"}{{{\color{jsonstring}{"}}}}{1},
}

\lstdefinelanguage{json}{
    basicstyle=\ttfamily\small,
    showstringspaces=false,
    breaklines=true,
    frame=lines,
    backgroundcolor=\color{metablue!10},  
    morestring=[b]",
    literate=
     *{0}{{{\color{jsonnumber}0}}}{1}
      {1}{{{\color{jsonnumber}1}}}{1}
      {2}{{{\color{jsonnumber}2}}}{1}
      {3}{{{\color{jsonnumber}3}}}{1}
      {4}{{{\color{jsonnumber}4}}}{1}
      {5}{{{\color{jsonnumber}5}}}{1}
      {6}{{{\color{jsonnumber}6}}}{1}
      {7}{{{\color{jsonnumber}7}}}{1}
      {8}{{{\color{jsonnumber}8}}}{1}
      {9}{{{\color{jsonnumber}9}}}{1}
      {:}{{{\color{jsonkey}{:}}}}{1}
      {,}{{{\color{jsonkey}{,}}}}{1}
      {"}{{{\color{jsonstring}{"}}}}{1},
}

\definecolor{orangebg}{HTML}{FFFAF0} 
\definecolor{orangefg}{HTML}{3E2723} 
\definecolor{orangekeyword}{HTML}{FF8C00} 
\definecolor{orangecyan}{HTML}{00BFFF} 
\definecolor{orangepurple}{HTML}{8A2BE2} 
\definecolor{orangegreen}{HTML}{32CD32} 
\definecolor{orangecomment}{HTML}{A52A2A} 
\definecolor{orangenumber}{HTML}{FF4500} 
\definecolor{bordercolor}{HTML}{FF4500} 

\lstdefinelanguage{custompython}{
    basicstyle=\ttfamily\small\color{orangefg}, 
    backgroundcolor=\color{metablue!10},  
    frame=single, 
    framerule=1pt, 
    framexleftmargin=2pt, 
    framexrightmargin=2pt, 
    framextopmargin=5pt, 
    framexbottommargin=5pt, 
    rulecolor=\color{metablue}, 
    showstringspaces=false, 
    breaklines=true, 
    keywordstyle=\color{orangekeyword}\bfseries, 
    stringstyle=\color{orangegreen},  
    commentstyle=\color{orangecomment}\itshape, 
    numberstyle=\tiny\color{orangenumber}, 
    identifierstyle=\color{orangefg}, 
    morekeywords={def, return, class, if, elif, else, for, while, import, from, as, pass, break, continue, lambda, try, except, raise}, 
    literate=
     *{0}{{{\color{orangecyan}0}}}{1}
      {1}{{{\color{orangecyan}1}}}{1}
      {2}{{{\color{orangecyan}2}}}{1}
      {3}{{{\color{orangecyan}3}}}{1}
      {4}{{{\color{orangecyan}4}}}{1}
      {5}{{{\color{orangecyan}5}}}{1}
      {6}{{{\color{orangecyan}6}}}{1}
      {7}{{{\color{orangecyan}7}}}{1}
      {8}{{{\color{orangecyan}8}}}{1}
      {9}{{{\color{orangecyan}9}}}{1}
      {:}{{{\color{orangepurple}{:}}}}{1}
      {,}{{{\color{orangepurple}{,}}}}{1}
      {"}{{{\color{orangegreen}{"}}}}{1},
}

\definecolor{marrsgreen}{RGB}{0, 140, 140}

\NewDocumentCommand{\mc}{ mO{} }{\textcolor{blue}{\textsuperscript{\textit{Mingchen}}}\textcolor{orange}{\textsf{\textbf{\small[#1]}}}}
\NewDocumentCommand{\wenyi}{ mO{} }{\textcolor{blue}{\textsuperscript{\textit{Wenyi}}}\textcolor{orange}{\textsf{\textbf{\small[#1]}}}}
\NewDocumentCommand{\dylan}{ mO{} }{\textcolor{blue}{\textsuperscript{\textit{Dylan}}}\textcolor{orange}{\textsf{\textbf{\small[#1]}}}}
\NewDocumentCommand{\cs}{ mO{} }{\textcolor{blue}{\textsuperscript{\textit{Changsheng}}}\textcolor{orange}{\textsf{\textbf{\small[#1]}}}}
\NewDocumentCommand{\yd}{ mO{} }{\textcolor{blue}{\textsuperscript{\textit{Yuandong}}}\textcolor{orange}{\textsf{\textbf{\small[#1]}}}}
\NewDocumentCommand{\yunyang}{ mO{} }{\textcolor{blue}{\textsuperscript{\textit{Yunyang}}}\textcolor{orange}{\textsf{\textbf{\small[#1]}}}}
\NewDocumentCommand{\zc}{ mO{} }{\textcolor{blue}{\textsuperscript{\textit{Zechun}}}\textcolor{orange}{\textsf{\textbf{\small[#1]}}}}
\NewDocumentCommand{\ernie}{ mO{} }{\textcolor{blue}{\textsuperscript{\textit{Ernie}}}\textcolor{orange}{\textsf{\textbf{\small[#1]}}}}
\NewDocumentCommand{\dmitrii}{ mO{} }{\textcolor{blue}{\textsuperscript{\textit{Dmitrii}}}\textcolor{orange}{\textsf{\textbf{\small[#1]}}}}
\NewDocumentCommand{\juergen}{ mO{} }{\textcolor{blue}{\textsuperscript{\textit{Juergen}}}\textcolor{orange}{\textsf{\textbf{\small[#1]}}}}
\NewDocumentCommand{\vikas}{ mO{} }{\textcolor{blue}{\textsuperscript{\textit{Vikas}}}\textcolor{orange}{\textsf{\textbf{\small[#1]}}}}
\NewDocumentCommand{\yangyang}{ mO{} }{\textcolor{blue}{\textsuperscript{\textit{Yangyang}}}\textcolor{orange}{\textsf{\textbf{\small[#1]}}}}
\NewDocumentCommand{\hz}{ mO{} }{\textcolor{blue}{\textsuperscript{\textit{Haozhe}}}\textcolor{orange}{\textsf{\textbf{\small[#1]}}}}

\title{
\begin{center}
Agent-as-a-Judge: \\Evaluate Agents with Agents
\end{center}
}

\author[1,2]{Mingchen Zhuge}
\author[1]{Changsheng Zhao}
\author[2]{Dylan R. Ashley}
\author[2]{Wenyi Wang}
\author[2]{Dmitrii Khizbullin}
\author[1]{Yunyang Xiong}
\author[1]{Zechun Liu}
\author[1]{Ernie Chang}
\author[1]{Raghuraman Krishnamoorthi}
\author[1]{Yuandong Tian}
\author[1]{Yangyang Shi}
\author[1]{Vikas Chandra}
\author[2]{Jürgen Schmidhuber}

\affiliation[1]{Meta AI}
\affiliation[2]{KAUST}

\abstract{
Contemporary evaluation techniques are inadequate for agentic systems. These approaches either focus exclusively on final outcomes---ignoring the step-by-step nature of agentic systems, or require excessive manual labour. To address this, we introduce the \textbf{Agent-as-a-Judge} framework, wherein agentic systems are used to evaluate agentic systems. This is an organic extension of the LLM-as-a-Judge framework, incorporating agentic features that enable intermediate feedback for the entire task-solving process.  We apply the Agent-as-a-Judge to the task of code generation. To overcome issues with existing benchmarks and provide a proof-of-concept testbed for Agent-as-a-Judge, we present \textbf{DevAI}, a new benchmark of 55 realistic automated AI development tasks. It includes rich manual annotations, like a total of 365 hierarchical user requirements. We benchmark three of the popular agentic systems using Agent-as-a-Judge and find it dramatically outperforms LLM-as-a-Judge and is as reliable as our human evaluation baseline. Altogether, we believe that Agent-as-a-Judge marks a concrete step forward for modern agentic systems---by providing rich and reliable reward signals necessary for dynamic and scalable self-improvement.
}

\date{\today}
\correspondence{\email{mingchen.zhuge@kaust.edu.sa}, \email{cszhao@meta.com}}
\metadata[Dataset]{\url{https://huggingface.co/devai-benchmark}}
\metadata[Project]{\url{https://github.com/metauto-ai/agent-as-a-judge}}
\addnote

\begin{document}

\maketitle

\section{Introduction} \label{sec:introduction}
\vspace{-5pt}

Recent years have seen multimodal agentic systems move from occasionally being able to solve small toy problems to being regularly deployed for challenging real-world problems (the dream of most AI research).
Yet, the current evaluation methods and the available benchmarks for agentic systems are struggling to keep up with these rapid advances, dramatically slowing true progress.

\begin{figure}[t]
\centering
\vspace{-10pt}
\includegraphics[width=0.95\linewidth]{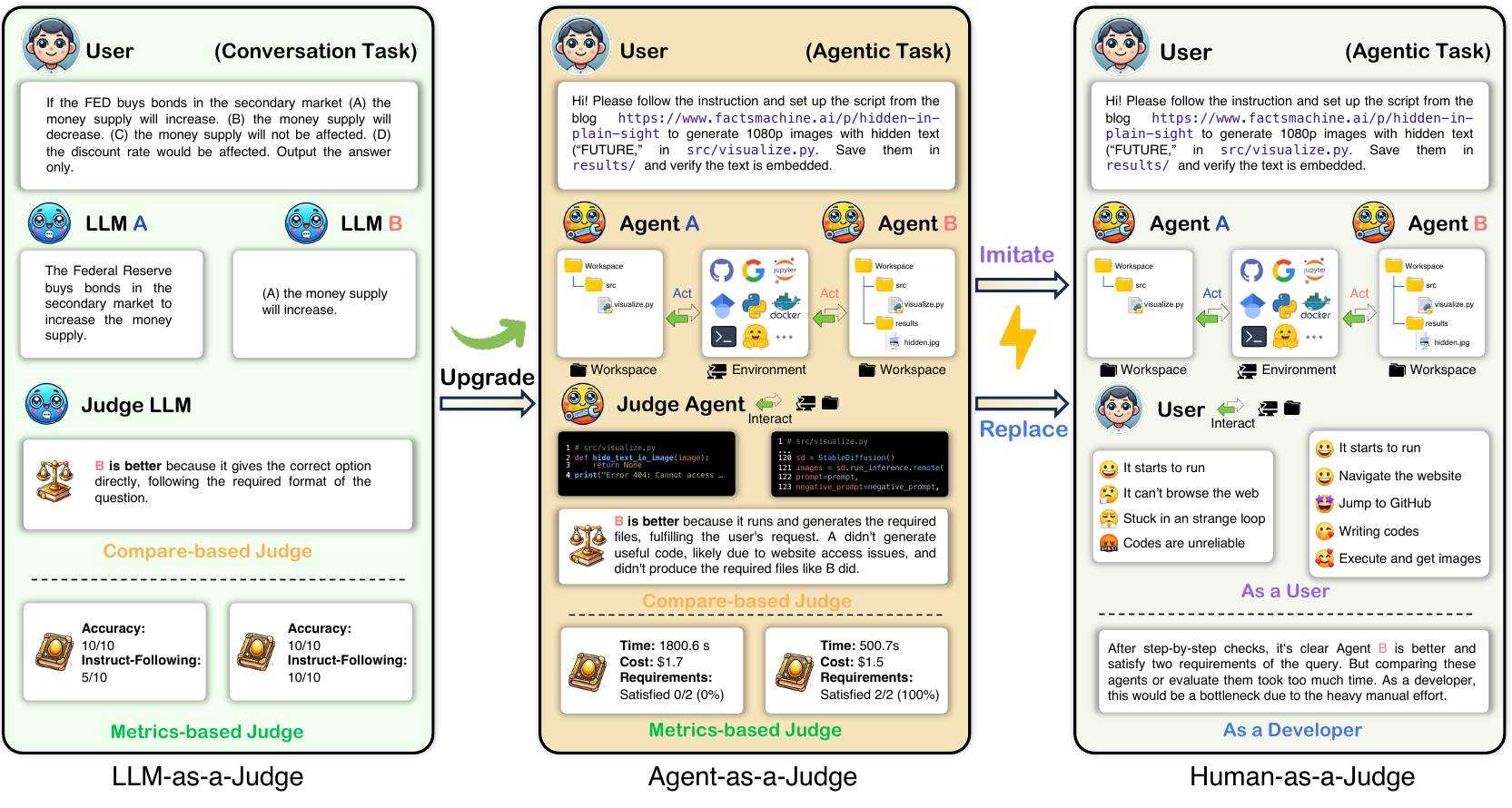}
\vspace{-5pt}
\caption{
We introduce the Agent-as-a-Judge framework wherein agentic systems are used to evaluate agentic systems.
We compare this to LLM-as-a-Judge, which uses LLMs to evaluate LLMs and for which Agent-as-a-Judge is a natural evolution, and Human-as-a-Judge, where skilled human labourers manually evaluate an agentic system.
}
\label{fig:ai_judges}
\end{figure}

We believe that the current issue with evaluating agentic systems stems from the lack of feedback during the intermediate task-solving stages for these nontraditional systems. Agentic systems think more like humans, often act step-by-step~\citep{wooldridge1999intelligent} and often host very human-like symbolic communications internally to solve problems~\citep{zhuge2023mindstorms}. 
And thus agentic systems should be evaluated like a human, with rich evaluative feedback which looks at the full thought and action trajectory; evaluating an agentic system in the traditional way is like evaluating a student using multiple-choice testing---a comparatively unreliable estimator~\citep{https://doi.org/10.1111/j.1467-8535.2010.01058.x}. 
For example, while SWE-Bench~\citep{yang2024swe} is widespread, its evaluation method, which relies solely on the final resolve rate for long-term automated repair tasks, does not effectively pinpoint what is happening within agentic systems that affects the resolve rate. 
On the other hand, performing a better evaluation with a human is prohibitively expensive. 
We instead propose that agentic systems should be used to evaluate agentic systems. 
Inspired by LLM-as-a-Judge~\citep{zheng2024judging,fu2023gptscore,chen2024mllm}, which uses LLMs to evaluate LLMs, we call this framework Agent-as-a-Judge, of which it is a key extension to the world of agentic systems (see Figure~\ref{fig:ai_judges}). 
It not only retains the cost-effectiveness of LLM-as-a-Judge but is also equipped with agentic features, allowing it to provide rich intermediate feedback throughout the entire process, as it acts as an agentic system.
We apply the Agent-as-a-Judge systems to the problem of evaluating code generating systems---one of the areas where agentic systems have looked the most promising recently.

In code generation, the development of benchmarks has also lagged behind the rapid advancement of agentic systems.
HumanEval~\citep{chen2021evaluating}, for example, focuses exclusively on algorithmic problems, while MBPP~\citep{austin2021program} deals with simple programming tasks.
Although they are useful for evaluating the basic skills of foundation models, neither of these two reflects the most practical challenges developers face. 
As a step away from this, SWE-Bench~\citep{jimenez2023swe} did introduce more realistic problems from GitHub, offering a fresh approach to evaluation, but still primarily focuses on automated repairs tasks development process. 
Concerningly, recent research shows that large language models (LLMs) can already solve over 27\% of the tasks in SWE-Bench without needing of advanced agentic systems~\citep{xia2024agentless}. 
Equally concerning, recent work has begun to introduce mechanisms designed specifically for the individual tasks in the SWE-Bench dataset, leading to a lack of real-world generalization and violating Goodhart's law: ``When a measure becomes a target, it ceases to be a good measure''~\citep{goodhart1976monetary}. 

To address the aforementioned issues with the current benchmarks in code generation, we introduce DevAI: the AI Developer Dataset, which contains $55$ real-world comprehensive AI app development tasks created by expert annotators.
We apply three leading open-source code-generating agentic frameworks to the tasks in DevAI: MetaGPT~\citep{hongmetagpt}, GPT-Pilot~\citep{GPTPilot2023}, and OpenHands~\citep{wang2024opendevin}.
We evaluate their performance using human judges (a painstaking process), LLM-as-a-Judge~\citep{zheng2024judging}, and our Agent-as-a-Judge framework.

Through human evaluation, we found that GPT-Pilot and OpenHands were each able to satisfy about $29\%$ of the task requirements in DevAI, but only one full task---showing that DevAI presents a good level of challenge to current systems.
When comparing our human judges with our automatic Agent-as-a-Judge framework, we found that Agent-as-a-Judge aligns more closely with the consensus of our human judges ($90\%$) as compared to LLM-as-a-Judge ($70\%$) in all cases tested. 
In addition, we find that it aligns more closely with this ensemble than the individual human evaluators do, suggesting that---not only is it suitable as a replacement for a human evaluator---but it could in fact be more useful than an average lone human evaluator.
In addition, considering the evaluation cost, Agent-as-a-Judge saves 97.72\% of the time and 97.64\% of the cost compared to involving three human experts.

In summary, the principal contributions of this work are:

\begin{itemize}[leftmargin=0.39cm]
    \item[$\bullet$]  We release the DevAI dataset, which consists of 55 comprehensive AI development tasks with accompanying tags, individual hierarchical requirements, and individual preferences. 
    \vspace{-2.5pt}
    \item[$\bullet$] We benchmark three top open-source code generation agentic frameworks in DevAI, providing a more comprehensive analysis than previous evaluations of them. 
    \vspace{-2.5pt}
    \item[$\bullet$] We introduce the general Agent-as-a-Judge concept, allowing agentic systems a fair and rich evaluation without the traditional costs associated with human involvement. 
    \vspace{-2.5pt}
    \item[$\bullet$] We demonstrate that an Agent-as-a-Judge outperforms an LLM-as-a-Judge and performs comparably to human evaluators in our proof-of-concept. 
\end{itemize}

\vspace{10pt}

\begin{tcolorbox}[colframe=metablue, colback=white, boxrule=0.5mm, width=\linewidth, boxsep=1mm, top=2mm, bottom=2mm]
    \textcolor{metablue}{\textbf{\lighticon\;Tips:} We provide a paper outline and the experimental design in \Cref{appendix:logic,appendix:experiment_design}.}
\end{tcolorbox}

\section{DevAI: A Dataset for Automated AI Development}\label{sec:dataset}

In this section, we introduce our new DevAI benchmark. We then evaluate three state-of-the-art code-generating agentic systems on this benchmark in \Cref{sec:human_evaluations} and present their basic statistics.

\subsection{Motivation}

\paragraph{\textbf{Background}}
The code generation domain is an area where agentic systems have seen significant industrial deployment over the past two years (e.g., Devin\footnote{\href{https://www.cognition.ai/blog/introducing-devin}{https://www.cognition.ai/blog/introducing-devin}} and Cursor\footnote{\href{https://www.cursor.com/}{https://www.cursor.com/}}).
However, in code generation, there isn't yet a benchmark that accurately reflects realistic user queries for developing complete AI systems. We believe this is because of the difficulty to evaluate such complex, real-world tasks.
For example, while many companies advertise their systems based on its performance on benchmarks such as SWE-Bench~\citep{yang2024swe} (for automated repair) or HumanEval~\citep{chen2021evaluating} (for algorithmic problems), these benchmarks cover only a small bit of an actual development process. Moreover, none of them accurately reflect the intermediate stages of development or provide sufficient reward signals for long-horizon development—similar issues are present in OpenAI's recent MLE-Bench~\citep{Chan2024MLEbenchEM}.  A benchmark that can evaluate the entire development process---ideally in a way that can help understand the degree to which current AI methods can reduce human labour---is missing.

\paragraph{\textbf{Topic}}
We chose automated AI development as our main topic. While AI and ML tasks are often more complex, they follow clear, standard procedures. For example, data processing typically comes first in an AI pipeline, and performance reporting goes at the end. We believe this topological nature can help better monitor the development process and provide useful signals to the agentic systems.

\paragraph{\textbf{Goals}}
An ideal benchmark should address critical issues in automated development by focusing on three key factors. First, it should reflect practical software scenarios, where tasks are often too complex for a single LLM, requiring human or agentic systems. Second, it should emphasize the development process, not just final outcomes (e.g., \texttt{pass@1} rates offer limited feedback and fail to highlight intermediate problems). Lastly, the evaluation should be computationally cost-effective and efficient, avoiding long training times or excessive manual oversight.

\begin{figure}[t]
    \centering
    \includegraphics[width=0.98\linewidth]{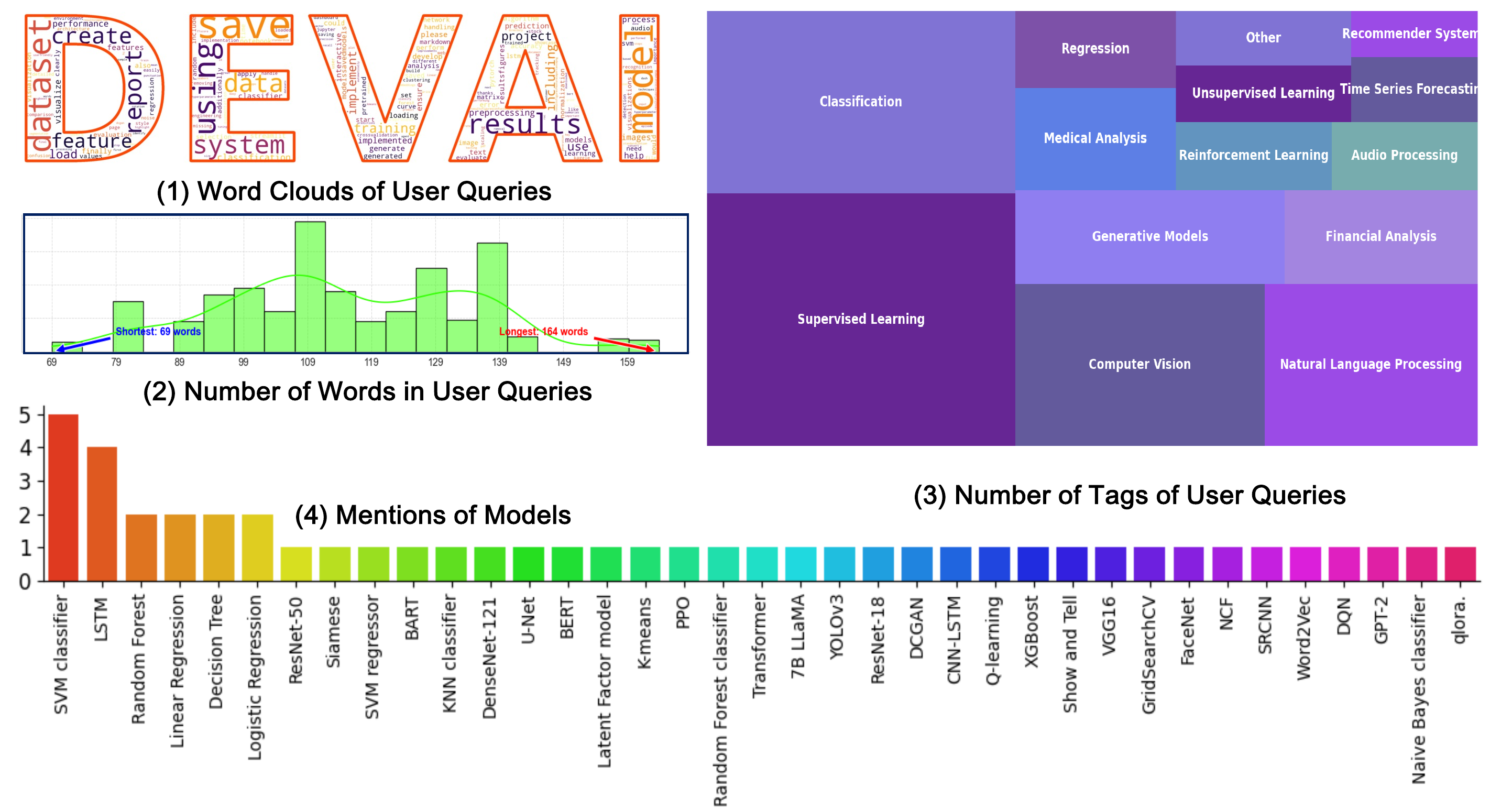}
    \vspace{-8pt}
    \caption{
    \textbf{Distribution of DevAI Tasks}
    (1) DevAI focuses on AI development tasks and so terms such as ``dataset,'' ``model,'' and ``results'' are particularly common in the queries.
    (2) The first 53 tasks in DevAI all have a one-paragraph query but of varying lengths (note that task 54 and 55 are excluded here as they are outliers, representing the longest and most complex tasks in the dataset).
    (3) Each task has one or more tags. The prevalence of supervised learning here reflects the fact that it dominates many machine learning applications.
    (4) SVM classifiers~\citep{cortes1995support} and LSTM models~\citep{hochreiter1997long} are two of the most widely used architectures---a fact reflected by DevAI.
    }
    \label{fig:user_query_overview}
\end{figure}

\begin{figure}[t!]
    \centering
    \begin{tcolorbox}[
        colframe=metablue,       
        colback=white,         
        coltitle=metablue,       
        title=\textbf{\textcolor{metablue}{Task 51: Devin\_AI\_Software\_Engineer\_Plants\_Secret\_Messages\_in\_Images}},
        fonttitle=\bfseries,   
        boxrule=0.5mm,         
        width=\textwidth,      
        halign=center,         
        colbacktitle=metablue!10 
    ]

\vspace{6pt}
    \begin{tcolorbox}[colframe=metablue, colback=metablue!3, sharp corners, boxrule=0.5mm, boxsep=5pt] 
        \begin{center}
            \textbf{\textcolor{metablue}{Query}}
        \end{center}
        \justifying  
        \emph{Hi! Please follow the instructions from the blog post \href{https://www.factsmachine.ai/p/hidden-in-plain-sight}{Hidden in Plain Sight} to set up the script for generating images with hidden text in \texttt{src/visualize.py}. Ensure the generated images are of 1080p resolution and saved in \texttt{results/}. Create control images embedding the text ``FUTURE" and save them in \texttt{results/}. Please manually verify that the hidden text is embedded in the images.}
    \end{tcolorbox}

    \vspace{10pt}
    
    \textbf{\textcolor{metablue}{Requirements}} 
    \begin{itemize}[left=0pt, itemsep=10pt, label={\scriptsize\textcolor{metablue}{$\blacksquare$}}]
        \item \textbf{\textcolor{metablue}{R0}} \\
        \textbf{Criteria:} \emph{Follow the instructions from the blog post \href{https://www.factsmachine.ai/p/hidden-in-plain-sight}{Hidden in Plain Sight} to set up the script for generating images with hidden text in \texttt{src/visualize.py}.} \\
        \textbf{Dependencies} \dependencyarrow\;\{\textbf{}\}
        
        \item \textbf{\textcolor{metablue}{R1}} \\
        \textbf{Criteria:} \emph{Ensure the generated images are of 1080p resolution and saved in \texttt{results/}.} \\
        \textbf{Dependencies} \dependencyarrow\;\{\textbf{\textcolor{metablue}{R0}}\}
        
        \item \textbf{\textcolor{metablue}{R2}} \\
        \textbf{Criteria:} \emph{Create control images embedding the text ``FUTURE" and save them in \texttt{results/}.} \\
        \textbf{Dependencies} \dependencyarrow\;\{\textbf{\textcolor{metablue}{R1}}\}
        
    \end{itemize}

    \vspace{10pt}

    \textbf{\textcolor{metablue}{Preferences (Optional)}} 
    \begin{itemize}[left=0pt, itemsep=10pt, label={\scriptsize\textcolor{metablue}{$\blacksquare$}}]
        \item \textbf{\textcolor{metablue}{P0}} \\
        \textbf{Criteria:} \emph{The system should be capable of learning and adapting to unfamiliar technologies and tools as required.}
        
        \item \textbf{\textcolor{metablue}{P1}} \\
        \textbf{Criteria:} \emph{After reviewing the blog post, ControlNet should successfully run on Modal to produce images with hidden messages for \texttt{FUTURE}.}
    \end{itemize}
     \vspace{10pt}
    \end{tcolorbox}
    \vspace{-5pt}
\caption{\textbf{A task example in DevAI}. This task is adapted from a real-world demo given at \url{https://www.cognition.ai/blog/introducing-devin}. As this example shows, task requirements in DevAI are structured as a Directed Acyclic Graph (DAG), with nodes representing individual requirements and directed edges showing dependencies. 
More examples are in \Cref{appendix:dataset_samples}.}
    \label{fig:task_51}
\end{figure}

\subsection{The DevAI Dataset}

Motivated by the ideas outlined above, we propose the DevAI dataset.
DevAI consists of a curated set of $55$ tasks, each defined by \textbf{(1)}~a plain text user query that describes an AI development task; \textbf{(2)}~a set of plain text requirements (for a total of $365$ requirements), each with a set of dependencies connecting them to other requirements; and \textbf{(3)} a set of preferences (for a total of $125$ preferences) which represent softer requirements.

DevAI is structured so that an agentic system starts by receiving a user query to begin development. The system is then evaluated on how well it meets the requirements, with preferences serving as optional, softer criteria. 
An example of one of the DevAI tasks can be seen in \Cref{fig:task_51}.

The tasks in DevAI are relatively small-scale but cover commonly used key development techniques. 
As shown in \Cref{fig:user_query_overview}, our tasks are tagged and cover a variety of key areas in AI: supervised learning, reinforcement learning, computer vision, natural language processing, generative models, and others.
Each of the tasks is a real-world problem that could be given to a research engineer, while simultaneously being relatively inexpensive computationally to run so as to reduce the cost of evaluating a method on this benchmark.
Details of the sample collection and human labeling process for DevAI are provided in \Cref{appendix:dataset}.

The requirements belonging to each task represent a milestone in the comprehensive development process and are arranged as a directed acyclic graph (similar to the work by \citet{he2021automl}), with requirements such as visualizing results depending on correct data loading and modeling.
This allows for more comprehensive non-sparse feedback than a binary success metric.
Furthermore, the inclusion of hierarchical requirements makes simple memorization an inadequate solution strategy, as completing the entire task requires agentic capabilities rather than relying solely on symbolic memorization, as is typical in foundation models.

\begin{table}[t]
  \centering
  \captionof{table}{
  \textbf{Preliminary Statistics of AI Developers.} 
   We compare three leading open-source code agents using metrics such as average cost, average time, and the average number of generated files.
  }
  \label{tab:ai_dev_comparison_1}
  \renewcommand\tabcolsep{1.8pt}
  \renewcommand\arraystretch{1.5} 
 \resizebox{1\columnwidth}{!}{%
  \begin{tabular}{lccc} 
    \Xhline{1.4pt}
    \rowcolor{metablue!10} 
    \multicolumn{1}{c}{\textbf{Metric}} & \textbf{MetaGPT}~\citep{hongmetagpt} & \textbf{GPT-Pilot}~\citep{GPTPilot2023} & \textbf{OpenHands}~\citep{wang2024opendevin} \\
    \Xhline{1.5pt}
    \rowcolor{gray!10} \multicolumn{4}{c}{\statisticsicon\;\textbf{Basic Statistics}} \\
    \cline{1-4} 
    \Xhline{1.5pt}
    \multicolumn{1}{l|}{\textcolor{gray}{Version} }        & \multicolumn{1}{c|}{\textcolor{gray}{Data Interpreter~\citep{hong2024data}}} & \multicolumn{1}{c|}{0.2.13} & \multicolumn{1}{c}{\textcolor{gray}{CodeAct v1.9~\citep{wang2024executable}}} \\
    \hdashline
      \multicolumn{1}{l|}{(1) Average Cost}         & \multicolumn{1}{c|}{\$1.19} & \multicolumn{1}{c|}{\$3.92} & \multicolumn{1}{c}{\$6.38} \\
    \multicolumn{1}{l|}{(2) Average Time}         & \multicolumn{1}{c|}{775.29s} & \multicolumn{1}{c|}{1622.38s} & \multicolumn{1}{c}{362.41s} \\
       \multicolumn{1}{l|}{(3) Average Input Tokens}         & \multicolumn{1}{c|}{152863} & \multicolumn{1}{c|}{606707} & \multicolumn{1}{c}{1252482} \\
     \multicolumn{1}{l|}{(4) Average Output Tokens}      & \multicolumn{1}{c|}{28546} & \multicolumn{1}{c|}{59707} & \multicolumn{1}{c}{8457} \\
      \hdashline
       \multicolumn{1}{l|}{(5) Average Saved Code Files}    & 
       \multicolumn{1}{c|}{0.42} & \multicolumn{1}{c|}{3.84} & \multicolumn{1}{c}{2.53} \\
     \multicolumn{1}{l|}{({6}) Average Saved Code Lines} & \multicolumn{1}{c|}{11.15} & \multicolumn{1}{c|}{273.33} & \multicolumn{1}{c}{96.56} \\ 
     \multicolumn{1}{l|}{(7) Average Saved Files}    & 
       \multicolumn{1}{c|}{4.42} & \multicolumn{1}{c|}{5.91} & \multicolumn{1}{c}{3.60} \\
    \Xhline{1.5pt}
  \end{tabular}
}
\end{table}

\subsection{Preliminary Benchmark}\label{exp:1a}

We first conduct experiments to collect development outcomes from different frameworks, which serve as baselines in the DevAI dataset. We test three of the most popular open-source frameworks (which we will refer to as ``\textbf{AI developers}''): MetaGPT~\citep{hongmetagpt}, GPT-Pilot~\citep{GPTPilot2023}, and OpenHands~\citep{wang2024opendevin}—all selected for their strong community acceptance (each having over 30,000 stars on GitHub). 

\paragraph{\textbf{Experiment Setup}}
 All of these three systems require a language model as a back-end engine, for which we use \texttt{gpt-4o-2024-05-13}, a state-of-the-art language model. These AI developers were given a time-limit of $1800$ seconds to solve each task and were forcefully halted if they exceeded this time limit (we imposed this constraint, which was visible to the AI developers, as detailed in \Cref{appendix:constraint}). 
We capture the outputs generated during the automated development process, including code, files, and other artifacts. Additionally, we record key decisions and actions made by the agentic systems through some custom instrumentation code, resulting in a development trajectory for each of the agentic systems.

\noindent \textbf{Analysis}\;\; 
The basic statistics are shown in \Cref{tab:ai_dev_comparison_1}. MetaGPT is the most cost-efficient ($1.19$ USD), while OpenHands is the most expensive ($6.38$ USD). In terms of development time, OpenHands completes tasks in an average of $362.41$s, while GPT-Pilot takes the longest at $1622.38$s. On average, a full evaluation on DevAI with one of these three took around $210.65$ USD and $14$ hours to perform.
While running, GPT-Pilot generates the most output tokens at $59707$ tokens, whereas OpenHands processed the most at $1 252 482$ tokens while producing the fewest at $8457$ tokens. This suggests that OpenHands's
internal communication is more complicated but is more parsimonious in its decisions. 

MetaGPT, while being the most cost-effective, generates fewer saved code files ($0.42$), suggesting it may be less inclined to save files. In contrast, GPT-Pilot generates the most saved files ($3.84$), reflecting a more prolific output. The difference in saved code lines, with GPT-Pilot saving $273.33$ lines versus MetaGPT’s $11.15$, underscores GPT-Pilot's extensive output. Meanwhile, OpenHands, despite handling larger inputs, seems less focused on executing code to generate files, as evidenced by its lower file output ($2.53$ saved files). These statistics align with real user experiences (as discussed in \Cref{appendix:user_experiences}).

\noindent \textbf{Evaluations}\;\; 
Note that the results in \Cref{tab:ai_dev_comparison_1} are not directly indicative of performance but provide valuable insights into the practical utility of DevAI and the performance of AI developers.
The generated workspaces (generated files, code, etc.) and trajectories are utilized in subsequent experiments to perform evaluations using Human-as-a-Judge (\cref{sec:human_evaluations}), LLM-as-a-Judge, and Agent-as-a-Judge (\cref{sec:aaaj}).

\section{Human-as-a-Judge: Manual Evaluation on DevAI}
\label{sec:human_evaluations}

To determine the pragmatic validity of DevAI and to accurately estimate the actual code-generating abilities of current state-of-the-art agentic systems, in this section, we run and then manually evaluate the application of three AI developer baselines to DevAI.
In \Cref{sec:aaaj}, we show how this evaluation can be automated.

\begin{table}[h]
  \centering
  \captionof{table}{
  \textbf{Human-as-a-Judge for AI Developers.} \textcolor{magenta}{(I)} and \textcolor{phthaloblue}{(D)} represent \textcolor{magenta}{independent} performance versus performance considering task \textcolor{phthaloblue}{dependencies}. \humanicon\; indicates multiple experts evolved, and \whiteboxicon\; means the evaluations use white-box testing (allowing access to the generated workspace, human-collected trajectories, and open-source codebases). The results were derived from expert judgments and deliberations (see \Cref{appendix:human_eval}).}
  \label{tab:ai_dev_comparison_2}
  \renewcommand\tabcolsep{2.1pt}
  \renewcommand\arraystretch{1.5} 
 \resizebox{1\columnwidth}{!}{%
  \begin{tabular}{lccc} 
    \Xhline{1.4pt}
    \rowcolor{metablue!10} 
    \multicolumn{1}{c}{\textbf{Metric}} & \textbf{MetaGPT}~\citep{hongmetagpt} & \textbf{GPT-Pilot}~\citep{GPTPilot2023} & \textbf{OpenHands}~\citep{wang2024opendevin} \\
    \Xhline{1.5pt}
    \rowcolor{gray!10} \multicolumn{4}{c}{\humanicon\;/ \whiteboxicon\;\textbf{Human-as-a-Judge}} \\
    \cline{1-4} 
    \Xhline{1.5pt}
     \multicolumn{1}{l|}{(A) Requirements Met \textcolor{magenta}{(I)}}     &  \multicolumn{1}{c|}{22.13\%} &  \multicolumn{1}{c|}{44.80\%} &  \multicolumn{1}{c}{42.89\%} \\
       \multicolumn{1}{l|}{(B) Requirements Met \textcolor{phthaloblue}{(D)}} &  \multicolumn{1}{c|}{6.55\%} &  \multicolumn{1}{c|}{28.96\%} &  \multicolumn{1}{c}{28.68\%} \\
     \multicolumn{1}{l|}{(C) Self-Termination} &  \multicolumn{1}{c|}{41.81\%} &  \multicolumn{1}{c|}{5.45\%} &  \multicolumn{1}{c}{54.54\%} \\
     \multicolumn{1}{l|}{(D) Task Solve Rate} &  \multicolumn{1}{c|}{0.00\%} &  \multicolumn{1}{c|}{1.81\%} &  \multicolumn{1}{c}{1.81\%} \\
    \Xhline{1.5pt}
  \end{tabular}
}
\end{table}

\subsection{Benchmark Baselines by Human-as-a-Judge}\label{exp:1b}

\noindent \textbf{Human Evaluation Setup}\;\; After obtaining the baseline executions and conducting basic statistical analysis, we have three expert human evaluators (referred to here by their anonymous names: \texttt{231a}, \texttt{38bb}, and \texttt{cn90}) review the outputs of AI developer baselines to assess whether each requirement was satisfied. 
We have two rounds of human evaluations.
To capture the bias inherent in typical human evaluation
(this is desirable to capture here as it represents a likely scenario in deployment),
in the first round, our evaluators first discussed the basic standards but were given minimal instructions.
The templates the evaluators were given for the evaluation and their self-reported post-hoc descriptions of how they resolved ambiguities are reported in \Cref{fig:evaluation_criteria} in \Cref{appendix:human_eval}.

After the initial round of human evaluations (which totaled an estimated total of $58$ human hours), we asked our evaluators to discuss and reach a consensus on their assessments (which took an estimated total of $28.5$ additional human hours). This consensus, achieved after long sessions of debate, was used as the final human evaluation result for each method.

\noindent \textbf{Performance Analysis}\;\; 
The results of this experiment are shown in \Cref{tab:ai_dev_comparison_2}.
We found that the two best-performing methods (GPT-Pilot and OpenHands) could satisfy about $29\%$ of the requirements (or around $44\%$ if prerequisites are ignored) but only on one task could they meet all the requirements.
This highlights that DevAI offers a considerable but appropriate level of challenge for current and future methods. 
Moreover, the fulfillment of intermediate requirements aligns with our expectations, as discussed in \Cref{sec:dataset}, that DevAI provides richer feedback by uncovering how agentic systems falter during the process instead of just focusing on a single performance metric at the end.

\subsection{Judging Human-as-a-Judge}\label{exp:2a}

\begin{figure}[h!]
    \centering
    \includegraphics[width=1\linewidth]{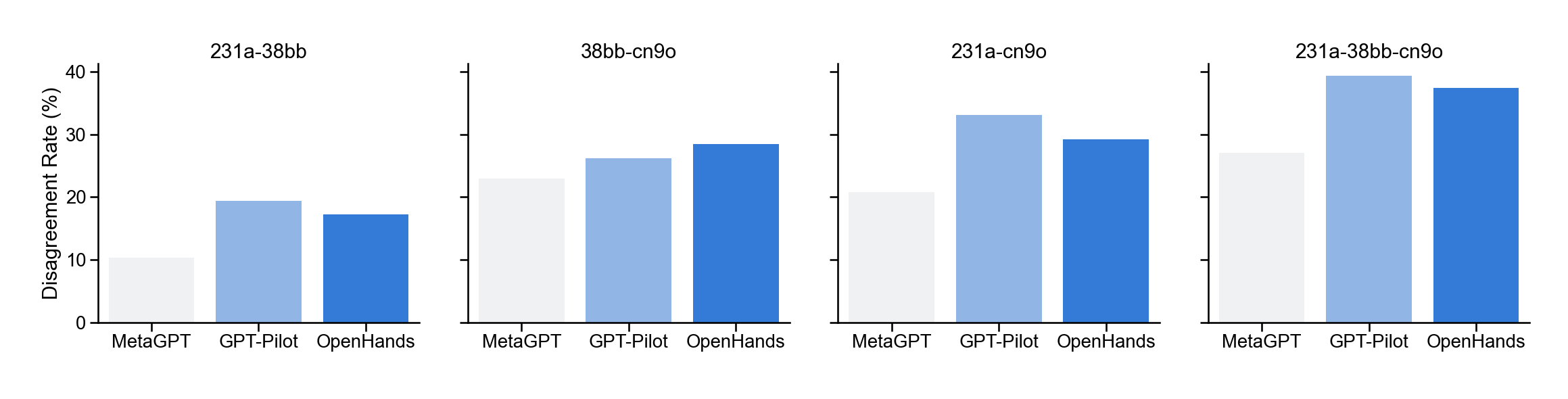}
    \vspace{-25pt}
    \caption{
    Between the three human evaluators, a large amount of disagreement was observed in their individual evaluations---highlighting the inherent unreliability of a single human evaluation. 
    }\label{fig:disagreement}
    \vspace{-2pt}
\end{figure}

\paragraph{\textbf{Disagreement Analysis}}
To analyze the presence of inductive bias and the reliability of the Human-as-a-Judge paradigm here, we calculate the disagreement rate between individual evaluators (shown in \Cref{fig:disagreement}). The results indicate that the disagreement rates between pairs of evaluators range from around $10\%$ to $30\%$.

Due to the complexity of a complete AI development task, which typically involves multiple steps with varying outcomes at each step, humans can easily make errors when critical information is missed, such as environment feedback indicating small but severe coding errors or bugs. Additionally, some disagreements are not necessarily incorrect but arise from differing perspectives on how ambiguity should be resolved.

\paragraph{\textbf{Error Analysis}}
As previously noted, the evaluators engaged in a round of debating after their initial evaluations until they reached a consensus on each requirement in each task (with the results of this consensus evaluation shown in \Cref{tab:ai_dev_comparison_2}). 

In our Human-as-a-Judge pipeline, evaluators could be convinced by evidence from others and acknowledge their judgment errors, adjusting their answers accordingly. This can be used to approximate individual errors. If the consensus evaluation more accurately predicts any extant ground truth, we would expect the majority vote from the individual evaluations to more closely approximate this than any single evaluation, due to the fundamental properties of ensemble classifiers (see \citet{hastie2009elements}).

\begin{wrapfigure}{r}{0.44\linewidth}
   \vspace{-26pt}
    \centering
    \includegraphics[width=1\linewidth]{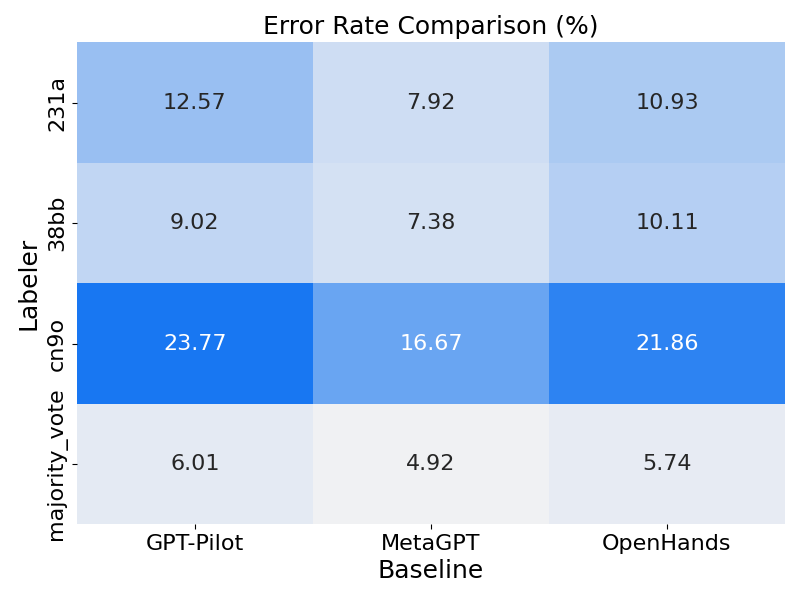}
    \vspace{-20pt}
    \caption{
        Mismatch between the individual evaluations and the consensus evaluation.
       In particular, the majority vote classifier showed the smallest deviation from the consensus evaluation.}
    \vspace{-20pt}
    \label{fig:error_rate}
\end{wrapfigure}

While the consensus evaluation may not represent the absolute ground truth (we acknowledge that some quantity of error likely would still exist even after this procedure), we expect the consensus evaluation to more accurately approximate any extant ground truth~\citep{clemen1989combining}. 
If this holds, the majority vote should align more closely with the consensus than with any individual evaluation. 
As shown in \Cref{fig:error_rate}, this is the case.

As seen in the results, although significant errors occur among all evaluators, the majority vote effectively corrects most of these errors. Notably, \texttt{cn9o} made the most errors (for example, $23.77\%$ in evaluating GPT-Pilot). After applying the majority vote from all three evaluators, the overall error rate dropped to $6.01\%$, demonstrating the inherent benefits of majority voting.

\paragraph{\textbf{Conclusion}} Human judgment errors are inevitable. To reduce them, we suggest two methods. First, like in this work, introduce a debate round after each judgment, where individuals present evidence and either persuade others or adjust their own opinions after discussion. This is particularly important when there are only a few evaluators, as majority voting with a small group can still lead to errors (around $5\%$ compared to consensus evaluation, as shown in \Cref{fig:error_rate}).
The second approach involves assembling a larger panel of experts (more is better when their accuracy exceeds $50\%$~\citep{grofman1983thirteen}), with over $5$ people recommended by \citet{hastie2005robust, larrick2006intuitions}, and relying on a majority vote. However, due to the high cost of engaging more experts and the fact that this is not always feasible in practice, we argue for the former.

\begin{table}[t!]
  \centering
  \captionof{table}{
  \textbf{AI Judges and Their Shift/Alignment with Human-as-a-Judge.} We compare the results of LLM-as-a-Judge and Agent-as-a-Judge with Human-as-a-Judge. \textcolor{magenta}{(I)} represents performance on independent tasks, while \textcolor{phthaloblue}{(D)} represents performance considering task \textcolor{phthaloblue}{dependencies}.  \textbf{Note:} \grayboxicon\;gray-box settings use carefully manually collected trajectory data (which is nearly inaccessible in practical situations, see \Cref{appendix:trajectory}). In contrast, \blackboxicon\;black-box setting doesn't need to access to such data. The red scores represent the absolute judge shift compared with Human-as-a-Judge (\emph{e.g.}, \textcolor{candypink}{2.74\%}).}\label{tab:ai_dev_comparison_alignment}
  \renewcommand\tabcolsep{4pt}
  \renewcommand\arraystretch{1.43} 
 \resizebox{1\columnwidth}{!}{%
  \begin{tabular}{lccc} 
    \Xhline{1.4pt}
    \rowcolor{metablue!10} 
    \multicolumn{1}{c}{\textbf{Metric}} & \textbf{MetaGPT}~\citep{hongmetagpt} & \textbf{GPT-Pilot}~\citep{GPTPilot2023} & \textbf{OpenHands}~\citep{wang2024opendevin} \\
    \Xhline{1.5pt}
    \rowcolor{gray!10}  \multicolumn{4}{c}{\textbf{\blackboxicon\; LLM-as-a-Judge}} \\
    \cline{1-4}  
    \Xhline{1.5pt}
     \multicolumn{1}{l|}{(a) Requirements Met \textcolor{magenta}{(I)} }     &  \multicolumn{1}{c|}{19.39\% (\textcolor{candypink}{2.74\%})} &  \multicolumn{1}{c|}{12.56\% (\textcolor{candypink}{32.24\%})} &  \multicolumn{1}{c}{11.47\% (\textcolor{candypink}{31.42\%})} \\
       \multicolumn{1}{l|}{(b) Requirements Met \textcolor{phthaloblue}{(D)}} &  \multicolumn{1}{c|}{1.63\% (\textcolor{candypink}{4.92\%})} &  \multicolumn{1}{c|}{4.09\% (\textcolor{candypink}{24.87\%})} &  \multicolumn{1}{c}{2.18\% (\textcolor{candypink}{26.50\%})}   \\ 
       \multicolumn{1}{l|}{(c) Task Solve Rate}  &  \multicolumn{1}{c|}{0.0\% (\textcolor{metablue}{0.0\%})} &  \multicolumn{1}{c|}{0.0\% (\textcolor{candypink}{1.81\%})} &  \multicolumn{1}{c}{0.0\% (\textcolor{candypink}{1.81\%})}   \\ 
       \hdashline
       \multicolumn{1}{l|}{\textcolor{metablue}{\textbf{Alignment Rate $\uparrow$}}}  &   \multicolumn{1}{c|}{\cellcolor{white!8}84.15\%} &  \multicolumn{1}{c|}{\cellcolor{white!0.5}65.30\%} &  \multicolumn{1}{c}{\cellcolor{white!0.5}60.38\%}  \\
    \Xhline{1.5pt}
    \rowcolor{gray!10}  \multicolumn{4}{c}{\textbf{\blackboxicon\; Agent-as-a-Judge}} \\
    \cline{1-4}  
    \Xhline{1.5pt}
     \multicolumn{1}{l|}{(I) Requirements Met \textcolor{magenta}{(I)} }     &  \multicolumn{1}{c|}{25.40\% (\textcolor{candypink}{3.26\%})} &  \multicolumn{1}{c|}{53.00\% (\textcolor{candypink}{8.20\%})} &  \multicolumn{1}{c}{42.62\% (\textcolor{candypink}{0.27\%})} \\
       \multicolumn{1}{l|}{(II) Requirements Met \textcolor{phthaloblue}{(D)}} &  \multicolumn{1}{c|}{5.73\% (\textcolor{candypink}{0.81\%})} &  \multicolumn{1}{c|}{39.89\% (\textcolor{candypink}{10.93\%})} &  \multicolumn{1}{c}{26.50\% (\textcolor{candypink}{2.17\%})}   \\
       \multicolumn{1}{l|}{(III) Task Solve Rate}  &  \multicolumn{1}{c|}{0.0\% (\textcolor{metablue}{0.0\%})} &  \multicolumn{1}{c|}{5.45\% (\textcolor{candypink}{3.64\%})} &  \multicolumn{1}{c}{1.81\% (\textcolor{metablue}{0.00\%})}   
      \\  
      \hdashline
       \multicolumn{1}{l|}{\textcolor{metablue}{\textbf{Alignment Rate $\uparrow$}}}  &   \multicolumn{1}{c|}{\cellcolor{white!10}88.52\%} &  \multicolumn{1}{c|}{\cellcolor{white!7}83.88\%} &  \multicolumn{1}{c}{\cellcolor{white!15}90.44\%}  \\
       \Xhline{1.5pt}
    \rowcolor{gray!10}  \multicolumn{4}{c}{\textbf{\grayboxicon \;LLM-as-a-Judge}} \\
    \cline{1-4}  
    \Xhline{1.5pt}
     \multicolumn{1}{l|}{(a) Requirements Met \textcolor{magenta}{(I)} }     &  \multicolumn{1}{c|}{28.68\% (\textcolor{candypink}{6.55\%})} &  \multicolumn{1}{c|}{38.79\% (\textcolor{candypink}{4.10\%})} &  \multicolumn{1}{c}{43.16\% (\textcolor{candypink}{0.27\%})} \\
       \multicolumn{1}{l|}{(b) Requirements Met \textcolor{phthaloblue}{(D)}} &  \multicolumn{1}{c|}{17.75\% (\textcolor{candypink}{11.20\%})} &  \multicolumn{1}{c|}{33.06\% (\textcolor{candypink}{4.10\%})} &  \multicolumn{1}{c}{32.24\% (\textcolor{candypink}{3.56\%})}   \\ 
       \multicolumn{1}{l|}{(c) Task Solve Rate}  &  \multicolumn{1}{c|}{1.81\% (\textcolor{candypink}{1.81\%})} &  \multicolumn{1}{c|}{3.63\% (\textcolor{candypink}{1.82\%})} &  \multicolumn{1}{c}{7.27\% (\textcolor{candypink}{5.46\%})}   \\ 
       \hdashline
       \multicolumn{1}{l|}{\textcolor{metablue}{\textbf{Alignment Rate $\uparrow$}}}  &   \multicolumn{1}{c|}{\cellcolor{white!2}68.86\%} &  \multicolumn{1}{c|}{\cellcolor{white!4}71.85\%} &  \multicolumn{1}{c}{\cellcolor{white!3}70.76\%}  \\

    \Xhline{1.5pt}
    \rowcolor{gray!10}  \multicolumn{4}{c}{\textbf{\grayboxicon \;Agent-as-a-Judge}} \\
    \cline{1-4}  
    \Xhline{1.5pt}
     \multicolumn{1}{l|}{(I) Requirements Met \textcolor{magenta}{(I)} }     &  \multicolumn{1}{c|}{23.49\% (\textcolor{candypink}{1.35\%})} &  \multicolumn{1}{c|}{46.44\% (\textcolor{candypink}{1.64\%})} &  \multicolumn{1}{c}{43.44\% (\textcolor{candypink}{0.54\%})} \\
       \multicolumn{1}{l|}{(II) Requirements Met \textcolor{phthaloblue}{(D)}} &  \multicolumn{1}{c|}{6.01\% (\textcolor{candypink}{0.54\%})} &  \multicolumn{1}{c|}{30.60\% (\textcolor{candypink}{1.64\%})} &  \multicolumn{1}{c}{28.14\% (\textcolor{candypink}{0.53\%})}   \\
       \multicolumn{1}{l|}{(III) Task Solve Rate}  &  \multicolumn{1}{c|}{0.0\% (\textcolor{metablue}{0.00\%})} &  \multicolumn{1}{c|}{5.45\% (\textcolor{candypink}{3.64\%})} &  \multicolumn{1}{c}{3.63\% (\textcolor{candypink}{1.82\%})}   
      \\  
      \hdashline
       \multicolumn{1}{l|}{\textcolor{metablue}{\textbf{Alignment Rate $\uparrow$}}}  &   \multicolumn{1}{c|}{\cellcolor{white!17}92.07\%} &  \multicolumn{1}{c|}{\cellcolor{white!10}86.61\%} &  \multicolumn{1}{c}{\cellcolor{white!15}90.16\%}  \\
       \Xhline{1.5pt}
    \rowcolor{gray!10} \multicolumn{4}{c}{\humanicon\;/ \whiteboxicon\;\textbf{Human-as-a-Judge}} \\
    \cline{1-4}
    \Xhline{1.5pt}
     \multicolumn{1}{l|}{\textcolor{metablue}{\textbf{Alignment Rate}} (38bb)} & \multicolumn{1}{c|}{\cellcolor{white!18}92.63\%} & \multicolumn{1}{c|}{\cellcolor{white!15}90.98\%} & \multicolumn{1}{c}{\cellcolor{white!13}89.89\%} \\
     \multicolumn{1}{l|}{\textcolor{metablue}{\textbf{Alignment Rate}} (cn9o)} & \multicolumn{1}{c|}{\cellcolor{white!7}83.33\%} & \multicolumn{1}{c|}{\cellcolor{white!5}76.23\%} & \multicolumn{1}{c}{\cellcolor{white!7}78.15\%} \\
     \multicolumn{1}{l|}{\textcolor{metablue}{\textbf{Alignment Rate}} (231a)} & \multicolumn{1}{c|}{\cellcolor{white!17}92.07\%} & \multicolumn{1}{c|}{\cellcolor{white!12}87.43\%} & \multicolumn{1}{c}{\cellcolor{white!13}89.07\%} \\
     \hdashline
   \rowcolor{gray!5}  \multicolumn{1}{l|}{\textcolor{gray!60}{\textbf{Average of individuals}}} & \multicolumn{1}{c|}{\textcolor{gray!60}{89.34\%}} & \multicolumn{1}{c|}{\textcolor{gray!60}{84.88\%}} & \multicolumn{1}{c}{\textcolor{gray!60}{85.70\%}} \\ 
   \rowcolor{gray!5}  \multicolumn{1}{l|}{\textcolor{gray!60}{\textbf{Best of individuals}}} & \multicolumn{1}{c|}{\textcolor{gray!60}{92.63\%}} & \multicolumn{1}{c|}{\textcolor{gray!60}{90.98\%}} & \multicolumn{1}{c}{\textcolor{gray!60}{89.89\%}} \\ 
     \hline
     \multicolumn{1}{l|}{\textcolor{metablue}{\textbf{Alignment Rate}} (Majority Vote)} & \multicolumn{1}{c|}{\cellcolor{white!22}95.08\%} & \multicolumn{1}{c|}{\cellcolor{white!19}93.98\%} & \multicolumn{1}{c}{\cellcolor{white!20}94.26\%} \\
  \Xhline{1.5pt}
  \end{tabular}
}
\end{table}

\section{Agent-as-a-Judge: Evaluating Agents with Agents}\label{sec:aaaj}

Human evaluation, while somewhat reliable, is time-consuming and requires substantial expertise. To address this, we propose the Agent-as-a-Judge framework.
If such an agentic system could evaluate like a human, it would reduce the need for human involvement and eliminate the trade-off between thoroughness and effort.

\subsection{Proof-of-Concept}
Based on our prior experiences with agent design and by imitating the human evaluation process, we initially designed eight modular, interacting components that form the foundation of our Proof-of-Concept for the Agent-as-a-Judge.

\begin{wrapfigure}{r}{0.5\linewidth}
   \vspace{-13pt}
    \centering
    \includegraphics[width=1\linewidth]{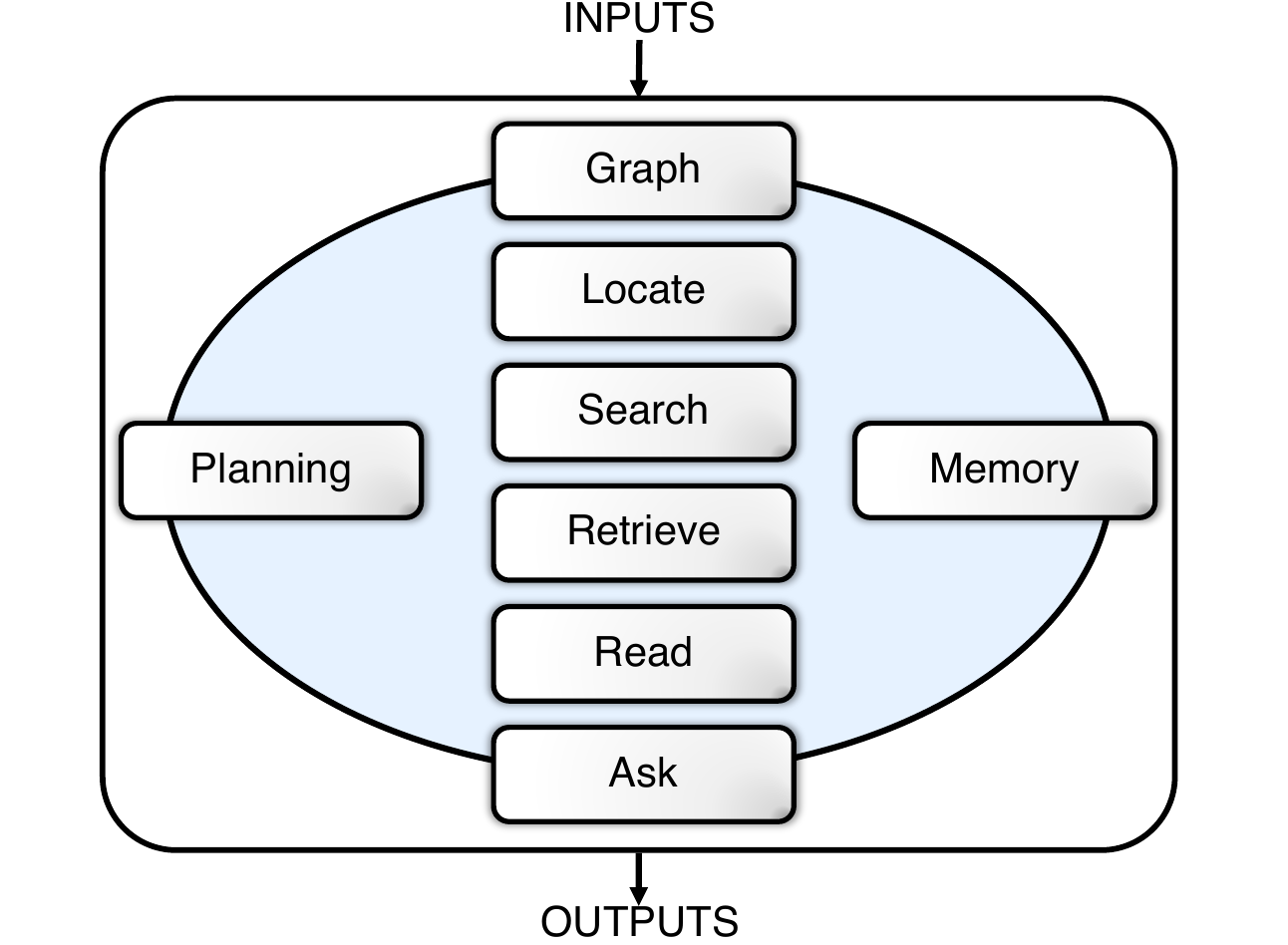}
    \vspace{-18pt}
    \caption{
       Initial diagram of Agent-as-a-Judge.}
    \vspace{-10pt}
    \label{fig:aaaj_poc}
\end{wrapfigure}

(1) The \texttt{graph} module constructs a graph that captures the entire structure of the project, including files, modules, and dependencies. It can also break down chunks of code into code snippets. 
(2) The \texttt{locate}  module identifies the specific folder or file referred to by a requirement. 
(3) The \texttt{read}  module goes beyond simple file parsing, supporting the reading and understanding of multimodal data across $33$ different formats, including code, images, videos and documents. This allows the agent to cross-reference various data streams and verify different kinds of requirement. 
(4) The \texttt{search}  module provides a contextual understanding of code and can quickly retrieve highly relevant code snippets, as well as the nuances behind them (e.g., hidden dependencies). 
(5) The \texttt{retrieve}  module extracts information from long texts, identifying relevant segments in trajectories. 
With context from the above, (6) the \texttt{ask}  module determines whether a given requirement is satisfied. 
(7) The \texttt{memory}  module stores historical judgment information, allowing the agent to build on past evaluations. 
Finally, (8) the \texttt{planning}  module plans the following actions, allowing the agent to strategize and sequence tasks based on the current state and the project goals.

Our initial design of the Agent-as-a-Judge, including all its components, is shown in \Cref{fig:aaaj_poc}, and the operational process of the Agent-as-a-Judge is illustrated in \Cref{fig:agent_as_a_judge}.

After conducting comprehensive ablation studies, we found that the modular combination of \text{(1)}, \text{(2)}, \text{(3)}, \text{(5)}, and \text{(6)} achieved the highest performance (see \Cref{appendix:ablation_aaaj}). A sample of the dynamic evidence collected by the Agent-as-a-Judge is shown in \Cref{appendix:judge_agent_evidence}.
We hypothesize this is because Agent-as-a-Judge needs high-quality factual information and is sensitive to noise. For example, while our design of the planning module introduces promising decision-making for future actions, the procedure is unstable. Initially, we hoped that historical information from the memory module would help to assess current requirements. However, it proved detrimental, as any errors in previous judgments could lead to a chain of errors, negatively affecting current decisions.  Besides, the current workspaces generated by developer agents, having only hundreds of lines of code, cannot fully benefit from the search module. 
The details of these findings are explained in \Cref{appendix:ablation_aaaj}.
Note that a perfect Agent-as-a-Judge is not the focus of this proof of concept, and thus, we leave the utilization of advanced agentic optimization methods for Agent-as-a-Judge, such as automated prompt optimization and workflow design~\citep{zhugegptswarm,hu2024automated}, for future work.

\subsection{Judging Agent-as-a-Judge and LLM-as-a-Judge}\label{exp:3a}

\noindent \textbf{Judge Shift}\;\; Judge Shift measures deviation from the Human-as-a-Judge consensus results, with lower values indicating a closer alignment. As shown in \cref{tab:ai_dev_comparison_alignment}, Agent-as-a-Judge consistently outperforms LLM-as-a-Judge across tasks, particularly those with task dependencies. For example, in Requirement \textcolor{magenta}{(I)}, Agent-as-a-Judge shows a Judge Shift as low as $0.27\%$, while LLM-as-a-Judge reaches $31.24\%$ for OpenHands. This underscores Agent-as-a-Judge's stability and suitability for meeting task requirements. Furthermore, in the gray-box setting, both Agent-as-a-Judge and LLM-as-a-Judge show even better results than their performance in the black-box setting.

\begin{wrapfigure}{r}{0.50\linewidth}
   \vspace{-10pt}
    \centering
    \includegraphics[width=1\linewidth]{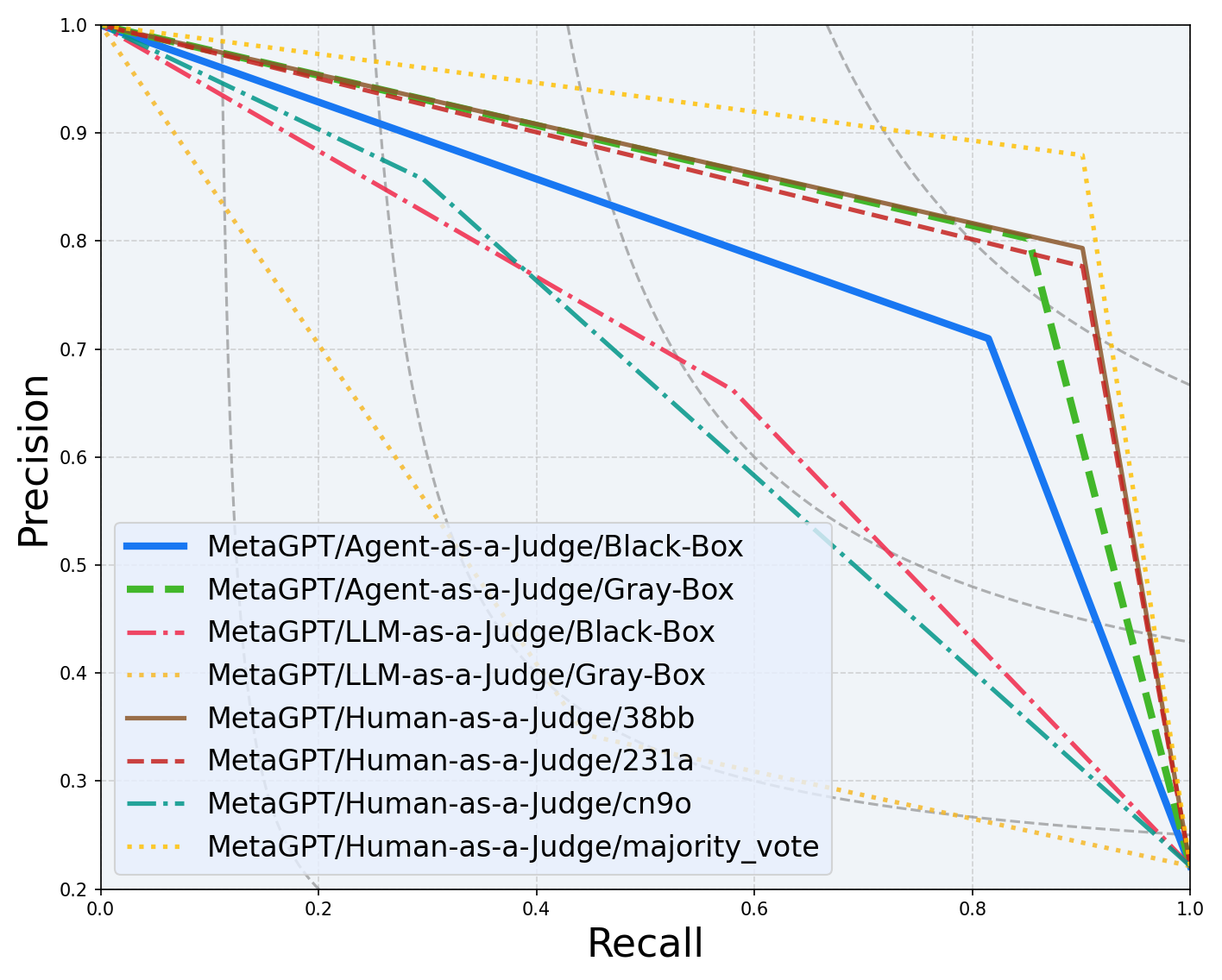}
    \vspace{-22pt}
    \caption{PR Curves comparing judge Methods.}
    \label{fig:pr_opendevin}
    \vspace{-10pt}
\end{wrapfigure}

\noindent \textbf{Alignment Rate}\;\; The Alignment Rate reflects how closely the AI Judges' evaluations align with human consensus across all $365$ requirements. It is defined as the percentage of requirement evaluations that are the same as the Human-as-a-Judge consensus evaluation. 
Compared to LLM-as-a-Judge, Agent-as-a-Judge consistently achieves a higher Alignment Rate, closely matching human judgments. For example, when evaluating OpenHands, Agent-as-a-Judge reaches $92.07\%$ and $90.44\%$, surpassing LLM-as-a-Judge's $70.76\%$ and $60.38\%$ in both gray-box and black-box settings. This shows that Agent-as-a-Judge produces more accurate and human-aligned evaluations, especially in complex scenarios.

\noindent \textbf{PR Curves}\;\; Judging developer agents is a class-imbalanced task, where meeting requirements is much rarer than failing. Metrics like judge shift and alignment rate can be misleading. For example, since MetaGPT rarely meets requirements, LLM-as-a-Judge easily identifies most cases as negative (achieving $84.15\%$ in the black-box setting). PR Curves offer a clearer performance measure by balancing precision and recall. Agent-as-a-Judge even outperforms any single human evaluator on OpenHands and aligns closest with majority voting. This shows that, in some cases, Agent-as-a-Judge can nearly replace human evaluators.

\subsection{Ablations For Agent-as-a-Judge}\label{exp:3b}

We conduct ablations to evaluate the impact of adding different components on Agent-as-a-Judge's performance. The components analyzed include \texttt{ask}, \texttt{graph}, \texttt{read}, \texttt{locate}, and \texttt{retrieve}. The component ablation study for Agent-as-a-Judge reveals key insights into the performance gains from adding specific functionalities. 

\begin{wraptable}{r}{0.55\textwidth}
  \centering
  \vspace{-15pt}
  \captionof{table}{
  \textbf{Component Ablation Studies for Agent-as-a-Judge.} We analyze the impact of adding various components (\texttt{ask}, \texttt{graph}, \texttt{read}, \texttt{locate}, and \texttt{retrieve}) on the performance of Agent-as-a-Judge for judging OpenHands. }
  \vspace{-3pt}
  \label{tab:agent_judge_ablation}
  \renewcommand\tabcolsep{2.1pt}
  \renewcommand\arraystretch{1.3} 
 \resizebox{0.55\textwidth}{!}{%
  \begin{tabular}{lccccc} 
    \Xhline{1.4pt}
    \rowcolor{metablue!20} 
    \multicolumn{1}{c}{\textbf{Metric}} & + \texttt{ask} & + \texttt{graph} & + \texttt{read} & + \texttt{locate} & + \texttt{retrieve} \\
    \Xhline{1.5pt}
    \rowcolor{gray!10}  \multicolumn{6}{c}{\textbf{Agent-as-a-Judge Performance}} \\
    \cline{1-6}  
    \Xhline{1.5pt}
     \multicolumn{1}{l|}{Alignment Rate}     &  \multicolumn{1}{c|}{65.03\%} &  \multicolumn{1}{c|}{75.95\%} &  \multicolumn{1}{c|}{82.24\%} & \multicolumn{1}{c|}{90.44\%} & \multicolumn{1}{c}{90.16\%} \\
\Xhline{1.5pt}
  \end{tabular}
}
\vspace{-13pt}
\end{wraptable}

With only the \texttt{ask} component, the agent achieves a $65.03\%$ alignment rate. Adding the \texttt{graph} component increases performance to $75.95\%$, as the agent can better understand the relationships between files.

The introduction of \texttt{read} further improves the alignment rate to $82.24\%$, reflecting the value of direct access to the contents of the file. Incorporating \texttt{locate} brings a substantial boost to $90.44\%$, as the agent can efficiently target files relevant to the requirements. However, adding \texttt{retrieve} does not provide a significant benefit in this case. In contrast, as shown in \Cref{tab:ai_dev_comparison_alignment}, the judgment of MetaGPT and GPT-Pilot indicates that \texttt{retrieve} is useful, as the trajectory provides additional valuable information.

\subsection{Cost Analysis}\label{exp:3c}

The Human-as-a-Judge took the three evaluators a self-reported total of $86.5$ hours. With a $15$ USD minimum wage (assuming this would buy a subject expert in AI), a full evaluation under DevAI would cost around $1297.50$ USD. In comparison, Agent-as-a-Judge cost only $30.58$ USD in API calls and took only $118.43$ minutes---$2.29\%$ of the cost and $2.36\%$ of the time of Human-as-a-Judge. LLM-as-a-Judge was faster at $10.99$ minutes, but due to the absence of intelligent context selection by the Agent-as-a-Judge's modules, it still cost $29.63$ USD.

\section{Related Work} \label{sec:related}

Agentic systems and their applications are highly active research areas with numerous recent works having a relation to this work.
This section details those works most relevant to ours.
We provide a treatment of the less relevant related works in \Cref{appendix:related_work}.

\noindent \textbf{AI Developers}\;\; 
AI in software development is growing fast~\citep{liu2024large}. AI-driven developers have been applied to directly imitate software companies~\citep{hongmetagpt,qian2024chatdev}, debug code~\citep{yang2024swe}, run data science methods~\citep{guo2024ds,hong2024data,li2024mlr,qiao2023taskweaver}, and even write academic papers~\citep{lu2024ai}.

\noindent \textbf{Benchmarks for AI developments}\;\; 
Benchmarks like MLAgentBench~\citep{huang2024mlagentbench}, ML-Bench~\citep{liu2023ml}, SUPER~\citep{bogin2024super}, DS-bench~\citep{jing2024dsbench}, and MLE-Bench~\citep{Chan2024MLEbenchEM}
all focus on benchmarking agentic systems using AI tasks.
However, DevAI distinguishes itself from all of these by focusing on realistic user queries that target a complete development cycle.
It further includes a more comprehensive evaluation with multiple hierarchical requirements and preferences for each task.
Comparatively, MLAgentBench~\citep{huang2024mlagentbench} for example, focuses on final performance for a limited set of well-known tasks, which risks overfitting and fails to assess a system’s generalization or adaptability.

\noindent \textbf{AI Judges}\;\;
Several works have looked at using AI systems as judges\footnote{Additionally, we were pleased to find that a recent industry blog (\url{https://www.cognition.ai/blog/evaluating-coding-agents}), published 3 weeks before our submission, shares very similar ideas and provides further evidence that the Agent-as-a-Judge could have practical applications in agent systems.}.
The work by \citet{chan2023chateval,zhao2024auto}, for example, extends LLM-as-a-Judge to have multiple LLMs in their evaluation process for conversational tasks.
Unlike Agent-as-a-Judge, they employ a trivial agentic system and apply it only to evaluate LLMs under traditional evaluation setups.
In contrast, \citep{lu2024llmscore} uses a single LLM-based evaluator but, unlike LLM-as-a-Judge, applies this to multimodal tasks rather than just for evaluating LLMs.
Less relevant are frameworks like those by~\citet{chen2024coder, arora2024masai, mundler2024code}, where intermediate signals are used during coding development.

\section{Discussion and Conclusion}

\paragraph{\textbf{Outlook 1: Intermediate Feedback for Agentic Self-Improvement}} 
A key power of the Agent-as-a-Judge, though not fully exploited here but nonetheless clear, is that it provides intermediate feedback that is essential for effective and efficient optimization~\citep{zhugegptswarm,pan2024autonomous}. 
For example,~\citet{agarwal2019learning} proposes to solve the sparse reward problem in reinforcement learning, by learning auxiliary reward functions that provide intermediate feedback.  
Perhaps the greatest strength of the Agent-as-a-Judge framework is that an agentic system can use it to identify and fix issues in its solutions to complex, multistage problems on the fly—something older, delayed-feedback methods did not permit.
By introducing Agent-as-a-Judge, we create the opportunity to build a process-supervised reward model (PRM) for improving agentic systems~\citep{lightman2023let}.

\paragraph{\textbf{Outlook 2:  Flywheel Effect Driven by Agent-as-a-Judge}}
The cycle of mutual improvement between the Agent-as-a-Judge and the evaluated agents, where both evolve together through iterative feedback, presents a promising direction. We hypothesize that an agentic version of a self-play system~\citep{zelikman2022star,chen2024self,wang2024self}, could viably emerge by using the Agent-as-a-Judge as a key mechanism. Furthermore, the ongoing interaction between the Agent-as-a-Judge and the evaluated agents has the potential to create a flywheel effect, where successive incremental improvements reinforce one another, leading to progressively greater optimization and enhanced performance over time~\citep{wang2022self}. This iterative process may also serve as a valuable complement to LLM reasoning data, help embedding agentic capabilities into foundation models~\citep{luo2024arena}.

\paragraph{\textbf{Conclusion}}
In this work, we introduced the Agent-as-a-Judge method to use agentic systems to evaluate agentic systems. We simultaneously released DevAI: a new benchmark that evaluates the code-generating ability of agentic systems on complete AI development tasks when used with Agent-as-a-Judge. We went on to show that Agent-as-a-Judge outperforms existing methods on this task and that it performs similarly to an ensemble of expert human evaluators. Altogether, we believe that the above opens the door for scaling up agentic far more than before.

\section*{Acknowledgements}

The authors thank Haozhe Liu, Piotr Piekos, Firas Laakom, Matteo Paltenghi for their suggestions or paper review. The research reported in this publication was supported by funding from the King Abdullah University of Science and Technology (KAUST) - Center of Excellence for Generative AI under award number 5940 and the SDAIA-KAUST Center of Excellence in Data Science and Artificial Intelligence.
\vspace{10pt}

\vspace{-0.3cm}

\bibliographystyle{plainnat}
\bibliography{related_work}

\newpage

\appendix

\appendix

\newpage

\section{Outline of this Paper}\label{appendix:logic}

\begin{figure}[h!]
    \centering
    \begin{tcolorbox}[
        colframe=metablue,       
        colback=white,         
        coltitle=metablue,       
        title=\textbf{\textcolor{metablue}{Paper: \emph{Agent-as-a-Judge: Evaluating Agents with Agents}}},
        fonttitle=\bfseries,   
        boxrule=0.5mm,         
        width=\textwidth,      
        halign=center,         
        colbacktitle=metablue!10 
    ]
    
    \vspace{5pt}
    
    \textbf{\textcolor{metablue}{Key Logic}} 
    \vspace{10pt}
    \begin{itemize}[left=0pt, itemsep=2pt, label={\scriptsize\textcolor{metablue}{$\blacksquare$}}]
        
        \item \textbf{\textcolor{metablue}{Step 1: Concept Proposal}} \\
        \textbf{Description:} We propose the \textbf{Agent-as-a-Judge} concept, an extension of the LLM-as-a-Judge framework, aimed at evaluating agentic systems using other agentic systems.

        \item \textbf{\textcolor{metablue}{Step 2: Dataset Creation}} \\
        \textbf{Description:} To address the lack of suitable datasets for evaluating agentic systems in automated AI development, we introduce \textbf{DevAI}, a new dataset consisting of 55 realistic AI code generation tasks. This also serves as a testbed for the Agent-as-a-Judge proof-of-concept.

        \item \textbf{\textcolor{metablue}{Step 3: Baseline Evaluation of Developer Agents \textcolor{orange}{(Experiment Level 1)}}} \\
        \textbf{Description:} In the first level of experiments, we select three popular open-source developer agents: \textbf{MetaGPT}, \textbf{GPT-Pilot}, and \textbf{OpenHands}. These agents are evaluated on the \textbf{DevAI} tasks to establish performance baselines.

        \item \textbf{\textcolor{metablue}{Step 4: Conducting Human-as-a-Judge Evaluation}} \\
        \textbf{Description:} We conduct a \textbf{Human-as-a-Judge} experiment, where three human experts assess the performance of the developer agents on the same DevAI tasks.

        \item \textbf{\textcolor{metablue}{Step 5: Human-as-a-Judge Analysis \textcolor{orange}{(Experiment Level 2)}}} \\
        \textbf{Description:} In the second level of experiments, we statistically analyze the results of Human-as-a-Judge evaluations, focusing on the costs of human labor and potential biases, highlighting the challenges of relying on human evaluation for complex tasks.

        \item \textbf{\textcolor{metablue}{Step 6: Agent-as-a-Judge Implementation}} \\
        \textbf{Description:} We design and implement the \textbf{Agent-as-a-Judge} proof-of-concept to evaluate code generation on the \textbf{DevAI} dataset. This system incorporates modules such as graph, search, read, and ask, providing multi-dimensional evaluation metrics.

        \item \textbf{\textcolor{metablue}{Step 7: Comparing AI Judge Systems \textcolor{orange}{(Experiment Level 3)}}} \\
        \textbf{Description:} In the third level of experiments, we compare three judgment systems: \textbf{Agent-as-a-Judge}, \textbf{LLM-as-a-Judge}, and \textbf{Human-as-a-Judge}, all applied to the same DevAI tasks. Our results show that \textbf{Agent-as-a-Judge} performs comparably to human evaluators and surpasses \textbf{LLM-as-a-Judge} in more complex reasoning and evaluation tasks.

    \end{itemize}

    \vspace{10pt}
    
    \textbf{\textcolor{metablue}{Future Directions}} 
    \vspace{10pt}
    \begin{itemize}[left=0pt, itemsep=2pt, label={\scriptsize\textcolor{metablue}{$\blacksquare$}}]
    
\item \textbf{\textcolor{metablue}{Direction 1: Intermediate Feedback for Agentic Self-Improvement}} \\
\textbf{Description:} \textbf{Agent-as-a-Judge} offers intermediate feedback, crucial for reinforcement learning, where rewards are sparse but vital for improvement. It also enables real-time issue identification and resolution in complex, multi-stage tasks, overcoming the limitations of delayed feedback.

\item \textbf{\textcolor{metablue}{Direction 2: Flywheel Effect Driven by Agent-as-a-Judge}} \\
\textbf{Description:} 
The iterative feedback cycle between the \textbf{Agent-as-a-Judge} and evaluated agents (such as \textbf{Developer Agents} here) could create a flywheel effect, where mutual improvements lead to progressively greater optimization. This dynamic could drive an agentic self-play system and complement LLM reasoning data to embed agentic features into foundation models.

    \end{itemize}

    \vspace{5pt}

    \end{tcolorbox}
\caption{\textbf{We Outline the Logical Flow of the Agent-as-a-Judge Framework}.  }
    \label{fig:logical_flow}
\end{figure}

\newpage
\section{Experiment Designs}\label{appendix:experiment_design}

This section outlines the experimental designs aimed at evaluating developer agents' performance, analyzing human evaluations, and comparing AI-based judging systems. The experiments are structured across three levels, as illustrated below.

\begin{center}
    \begin{tikzpicture}
        \node[anchor=south west, inner sep=0] (image) at (0,0) {\includegraphics[width=0.9\linewidth]{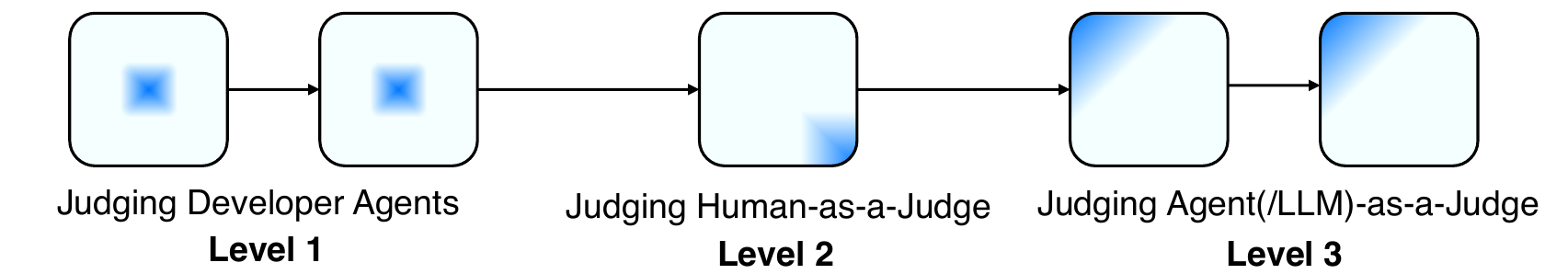}};
        \begin{scope}[x={(image.south east)},y={(image.north west)}]
            \node at (0.096, 0.68) {\scriptsize \Cref{exp:1a}};
            \node at (0.255, 0.68) {\scriptsize \Cref{exp:1b}};
            \node at (0.497, 0.68) {\scriptsize \Cref{exp:2a}};
            \node at (0.735, 0.68) {\scriptsize \Cref{exp:3a}};
            \node at (0.890, 0.68) {\scriptsize \Cref{exp:3b}};
        \end{scope}
    \end{tikzpicture}
\end{center}

\subsection{Summary of Experiments}

The experiments are categorized into three levels as follows:

\begin{itemize}[leftmargin=0.5cm] 
    \item[] \textcolor{metablue}{\textbf{Level 1}}
    \begin{itemize}[leftmargin=0.8cm] 
        \item[\textcolor{metablue}{\textleaf}] \emph{Experiment 1a:} Basic performance statistics for developer agents (\Cref{exp:1a}) 
        \item[\textcolor{metablue}{\textleaf}] \emph{Experiment 1b:} Human evaluations of developer agents (\Cref{exp:1b})
    \end{itemize}
    
    \item[] \textcolor{metablue}{\textbf{Level 2}} 
    \begin{itemize}[leftmargin=0.8cm]
        \item[\textcolor{metablue}{\textleaf}] \emph{Experiment 2a:} Error analysis of human evaluations (\Cref{exp:2a})
    \end{itemize}
    
    \item[] \textcolor{metablue}{\textbf{Level 3}} 
    \begin{itemize}[leftmargin=0.8cm]
        \item[\textcolor{metablue}{\textleaf}] \emph{Experiment 3a:} AI judge baselines (\Cref{exp:3a})
        \item[\textcolor{metablue}{\textleaf}] \emph{Experiment 3b:} Ablation studies for Agent-as-a-Judge (\Cref{exp:3b})
    \end{itemize}
\end{itemize}

\subsection{Judges and Subjects of Evaluation}

The following table summarizes the judge and the subject being evaluated in each experiment:

\begin{table}[h!]
  \centering
  \label{tab:judges_subjects_overview}
 \resizebox{1\columnwidth}{!}{%
  \renewcommand\tabcolsep{40pt}
  \renewcommand\arraystretch{1.3} 
  \begin{tabular}{llc} 
    \Xhline{1.2pt}
    \rowcolor{metablue!15} 
    \textbf{Experiment} & \textbf{Who is the Judge?} & \textbf{Who is being Judged?} \\
    \Xhline{1.2pt}
    \Cref{exp:1a} & \emph{Human} & \emph{Developer Agents} \\
    \hdashline
    \Cref{exp:1b} & \emph{Human} & \emph{Developer Agents} \\
    \Xhline{1.2pt}
    \Cref{exp:2a} & \emph{Human} & \emph{Human} \\
    \Xhline{1.2pt}
    \Cref{exp:3a} & (1) \emph{LLM-as-a-Judge} & (1) \emph{Developer Agents} \\
                  & (2) \emph{Agent-as-a-Judge} & (2) \emph{Developer Agents} \\
                  & (3) \emph{Human} & (3) \emph{LLM-as-a-Judge} \\
                  & (4) \emph{Human} & (4) \emph{Agent-as-a-Judge} \\
    \hdashline
    \Cref{exp:3b} & \emph{Human} & \emph{Agent-as-a-Judge} \\
    \Xhline{1.2pt}
  \end{tabular}
}
\end{table}

\newpage
\section{Agent-as-a-Judge Pipeline}\label{appendix:aaaj_pipeline}

\begin{figure}[h]
\centering
\includegraphics[width=0.88\linewidth]{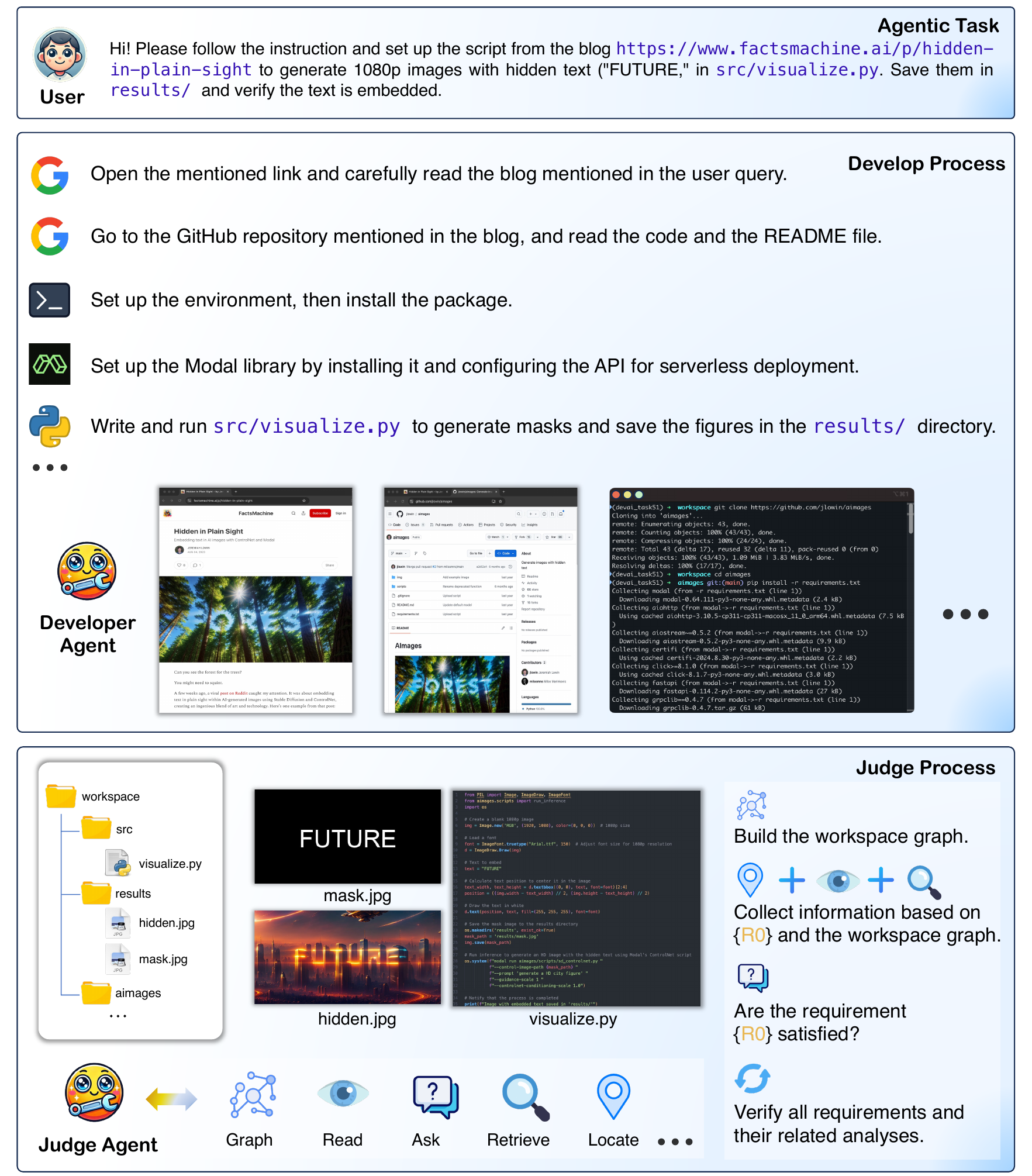}
\caption{
The pipelines of developer agents and judge agent. Some materials in this figure are from original blog (\url{https://www.factsmachine.ai/p/hidden-in-plain-sight}). 
}
\label{fig:agent_as_a_judge}
\end{figure}

\newpage

\section{Extend Related Work}\label{appendix:related_work}

Our main paper includes mostly related works of \textbf{AI developers}, \textbf{Benchmarks for AI developments}, and \textbf{AI judges}. 
However, the following works contribute significantly to the community and also relate to this work. We record this work as additional related work.

\paragraph{\textbf{LLM-based Autonomous Agents}} Recent developments in LLM-based agents have expanded their capabilities beyond simple task execution to more autonomous problem-solving and decision-making. AutoGPT~\citep{gravitas2023auto} and LangChain~\citep{Chase_LangChain_2022} provide frameworks for single-agent systems that leverage external tools for more complex tasks. Similarly, research such as CAMEL~\citep{li2023camel}, MetaGPT~\citep{hongmetagpt}, NLSOM~\citep{zhuge2023mindstorms}, AutoGen~\citep{wu2023autogen} focus on role-based multi-agent communication, improving collaboration among agents. However, the challenge of maintaining coherence in agents' dialogue and preventing hallucination remains prominent~\citep{du2024multi,zhou2023agents}. Besides, Agent-trust~\citep{xie2024can} examines if LLM agents, like GPT-4, can simulate human trust behaviors in Trust Games, showing that they can align with human behavior.

Most recently, using graphs to build agents has gained prominence. Earlier work like GPTSwarm~\citep{zhugegptswarm} and LangGraph~\citep{langgraph} proposed using nodes to represent operations and edges to represent the connections between them. In GPTSwarm, multiple agents represented as subgraphs in a graph are connected by optimizable edges, and reinforcement learning is employed to optimize the edges. Following this approach, several agent frameworks have incorporated graphs into their designs~\citep{hong2024data,zhou2024symbolic,qian2024scaling}. Additionally, various optimization methods have been developed to enhance agent performance further~\citep{wu2024stateflow,song2024adaptive,hu2024automated}. In practical applications, many studies focus on understanding and interacting with GUIs~\citep{wang2024mobile, chen2024gui, yang2023appagent, xu2024crab, tan2024towards, wu2024copilot}. For code generation agents~\citep{jin2024llms}, current research mainly emphasizes automated repair~\citep{yang2024swe, phan2024hyperagent, tao2024magis}, computational modular design~\citep{khattab2023dspy, cheng2024trace}, and automated development~\citep{tufano2024autodev, huang2023agentcoder}. Among these, open-sourced frameworks like OpenHands~\citep{wang2024opendevin} have gained popularity due to their strong user experience. Moreover, scientific discovery~\citep{jansen2024discoveryworld, lu2024ai} and ML agents~\citep{yang2024matplotagent} are also receiving increased attention.

\paragraph{\textbf{LLM-as-a-Judge}} In the domain of AI evaluation and judgment, frameworks~\citep{zheng2024judging,fu2023gptscore} have pioneered the use of LLMs to assess conversational agents, demonstrating how LLMs can evaluate dialogue quality and consistency. 
LLM-as-a-judge has also expanded into the multimodal domain, providing clear visual-language feedback~\citep{chen2024mllm,xiong2024llava}.
ICE-Score~\citep{zhuo2023ice} improves on metrics like CodeBERTScore~\citep{zhou2023codebertscore} and G-EVAL~\cite{liu2023g} by better aligning with human preferences and functional correctness.
Expanding beyond dialogue, LLMs like CodeR~\citep{chen2024coder} and MASAI~\citep{arora2024masai} apply similar judging principles to the code validation process, where AI systems autonomously evaluate and verify computer programs. Our work builds on these advancements by exploring how LLMs can perform more nuanced judgment tasks, further investigating their potential in decision-making across various domains. Recent research also focuses on judging LLM-as-a-Judges~\citep{chen2024humans, bavaresco2024llms, thakur2024judging, dong2024can, shi2024judging, raina2024llm}.

\paragraph{\textbf{Coding Benchmarks}}

Recent advances in code generation have led to the innovation of various benchmarks to evaluate model performance~\citep{liu2024large}. Early benchmarks, such as MBPP~\citep{austin2021program}, HumanEval~\citep{chen2021evaluating}, and MultiPL-E~\citep{cassano2023multipl}, focus primarily on generating simple functions. While these benchmarks are useful for evaluating the correctness of generated code, they are limited in complexity and do not fully represent the challenges encountered in real-world software development.

As the field progressed, newer benchmarks began to focus on more complex and realistic tasks. APPS~\citep{hendrycks2021measuring},
CodeContests~\citep{li2022competition}, and LiveCodeBench~\citep{jain2024livecodebench} moved toward competitive programming challenges that involve advanced algorithms and data structures. These tasks are more representative of problems encountered in coding competitions and help push models toward more sophisticated problem-solving.
DS-1000~\citep{lai2023ds} was introduced to assess the skills of models with data science libraries, evaluating their ability to use APIs and execute complex data analysis workflows. Meanwhile, AgentBench~\citep{liu2023agentbench} focuses on testing reasoning and decision-making abilities in interactive environments, highlighting differences in performance between commercial and open-source models. 

To address real-world programming needs beyond code generation, specialized benchmarks have been created to evaluate tasks such as debugging, refactoring, and code navigation.  BigCodeBench~\citep{zhuo2024bigcodebench} has advanced the evaluation of LLMs by testing their ability to solve complex, real-world programming tasks beyond traditional algorithmic challenges. CANITEDIT~\citep{cassano2024editevaluatingabilitylarge}, DebugBench~\citep{tian2024debugbench}, and FixEval~\citep{haque2023fixeval} evaluate the ability of a model to edit and improve existing code. Additionally, benchmarks such as SWE-Bench~\citep{jimenez2023swe} focus on resolving issues in GitHub repositories, simulating practical software development scenarios. Finally, benchmarks such as RepoBench~\citep{liu2023repobench} and RepoEval~\citep{zhang2023repocoder} delve into the evaluation of models in large-scale, multifile codebases.
These benchmarks measure the ability of language models to understand the structure of repositories and solve problems within more complex, collaborative environments.

There are also tasks that assess tool-use skills requiring programming knowledge, though not directly focused on programming. CRAB~\citep{xu2024crab} sets a new standard for benchmarking autonomous agents by supporting cross-environment tasks and offering a more comprehensive evaluation framework. Similarly, WebArena-serious~\citep{zhou2023webarena,koh2024visualwebarena} introduces a realistic benchmark for evaluating multimodal agents on complex, real-world web tasks, addressing the limitations of text-only agents by integrating both visual and textual inputs across diverse domains.

\newpage

\section{The Procedures of Creating DevAI Dataset}\label{appendix:dataset}

\subsection{Manually draft user queries}
Given the execution cost of the developer agents, we collect small-scale AI tasks to ensure the practical applicability of our benchmark.
Since these tasks are small-scale and well studied, which are easy to overfit in terms of task performance metrics, unlike previous benchmarks (e.g., ~\citep{huang2024mlagentbench}), we do not evaluate task performance as the development performance measure. 
Instead, we prioritize the step-by-step task-solving ability, which is essential for real-world development. 
Our quires are specifically designed to require the development agents to understand user intentions, solve the task in multiple steps, and adapt to unexpected step outcomes.
This approach also makes our benchmark user-friendly, transparent, and better reflects real-world deployment situations.
To enable effective evaluation, our queries present a specific development file structure for the developer agents to follow.
To ensure that the developer agents save the files to be evaluated in the workspace, we develop constraint prompts added to the query to form an extended query. The constraint prompts guide the developer agents to save and execute the source codes, which are in line with the real-world development standard. See Appendix~\ref{appendix:constraint} for our constraint prompts.

\subsection{Set Judging Criteria}
To make the evaluation of the developer agent precise, we assign to each task query a list of requirements as task milestones. 
The requirements are chosen so that satisfying all the requirements is a necessary condition to consider the task to be solved successively.
Since our tasks are AI-centric, our queries target essential elements of AI development, including data processing, AI method, presentation of evaluation metrics, visualization, and human-computer interaction, covering the key areas that matter most in real-world scenarios.
On the other hand, breaking down tasks into individual requirements also reflects the multi-step nature of code development.
Importantly, to avoid ambiguity, we set the requirements to be explicit, binary, and straightforward to evaluate. To include other human predispositions, we include a list of preferences per task that covers subjective, ambiguous, or non-explicitly stated characteristics.

\subsection{Building Dependency Among Requirements}
To enhance the realism of our benchmark, we analyzed the dependencies among requirements. Over the past decades, methodologies such as the KDD Process~\citep{fayyad1996data} and CRISP-DM~\citep{wirth2000crisp} have guided ML/AI development, establishing foundational frameworks that have been further refined with the rise of AutoML~\citep{he2021automl}. 

Inspired by these methodologies, we identified a seven-step process for automated AI development tasks. This process includes critical stages such as data preprocessing, feature engineering, model selection, and hyperparameter tuning, along with essential post-development activities like metrics recording, report generation, and the development of interactive user applications. This structured approach allows us to evaluate the agent's ability to manage task dependencies and effectively navigate complex, real-world scenarios.

\subsection{Refine the dataset}
Manual refinements were necessary to ensure the accuracy and clarity of DevAI.
We perform two rounds of comprehensive review and edits on DevAI, each round being done by a different participant. 
During these manual refinements, we focus on the logical consistency of our queries and requirements, the correctness and ambiguity of the language, and the applicability of the content to the task domain.
We identified a moderate number of errors in our dataset during the review.

\subsection{Analyse the dataset}
We categorized each requirement based on its focus, whether it was related to the data set, the machine learning method, visualization, metrics, HCI, or data processing. While this approach provides a useful framework, it is important to recognize that these categories might overlap or miss certain nuances. Similarly, preferences were classified by how strongly they appeared in the query text, ranging from those inferred by common sense to those explicitly stated. Although this helps to organize preferences, it is worth noting that such classifications can be subjective and may not fully capture the importance of each preference in practical applications. By structuring the requirements and preferences this way, we aim to enhance the evaluation process, though flexibility and context awareness remain crucial for truly robust assessments.

\subsection{Auxiliary Information}
Some of the tasks require the download of a Kaggle~\cite{} data set, where a Kaggle credential is needed.
Our constraint~\ref{appendix:constraint} requires an ``\texttt{is\_kaggle\_api\_needed}" tag to determine whether the credential is needed to be included in the extended query.
We further mark each task with some tags describing the AI sub-fields related to the task, including computer vision, supervised learning, reinforcement learning, natural language processing, etc., as well as two ``\texttt{is\_training\_needed}" and ``\texttt{is\_web\_navigation\_needed}" tags as auxiliary information. 
We also categorize each requirement into one of the following: (1) \texttt{dataset or environment}, (2) \texttt{data preprocessing and postprocessing}, (3) \texttt{machine learning method}, (4) \texttt{save trained model}, (4) \texttt{performance metrics}, (5) \texttt{human computer interaction}, (6) \texttt{visualization}, and (7) \texttt{other}, reflecting the nature of the requirement.

\subsection{A Json Format of Our Sample}

Here, we provide a sample of the DevAI with its json format. We also provide more samples in \Cref{appendix:dataset_samples}.

\begin{lstlisting}[language=json]
{
    "name": "25_Speech_Emotion_Recognition_CNN_LSTM_RAVDESS_DL",
    "query": "I am seeking a speech emotion recognition project using a CNN-LSTM model with the RAVDESS dataset, which should be downloaded from Kaggle or [this Hugging Face link](https://huggingface.co/datasets/xbgoose/ravdess). The project should load the dataset and perform robust audio preprocessing (noise removal and normalization) and MFCC feature extraction, implemented in `src/data_loader.py`. The CNN-LSTM model should be implemented in 'src/model.py'. Recognition accuracy should be saved in `results/metrics/recognition_accuracy.txt`, and a confusion matrix should be generated and saved as `results/figures/confusion_matrix.png`. Additionally, a user-friendly local API should be created using Flask to allow users to upload audio files and receive emotion recognition results, with the implementation included in `src/hci.py`.",
    "tags": [
        "Audio Processing",
        "Classification"
    ],
    "requirements": [
        {
            "requirement_id": 0,
            "prerequisites": [],
            "criteria": "The \"RAVDESS\" dataset is loaded in `src/data_loader.py`, which is downloaded from Kaggle or [this Hugging Face link](https://huggingface.co/datasets/xbgoose/ravdess).",
            "category": "Dataset or Environment",
            "satisfied": null
        },
        {
            "requirement_id": 1,
            "prerequisites": [
                0
            ],
            "criteria": "Audio preprocessing, including noise removal and normalization, is implemented in `src/data_loader.py`.",
            "category": "Data preprocessing and postprocessing",
            "satisfied": null
        },
        {
            "requirement_id": 2,
            "prerequisites": [
                0,
                1
            ],
            "criteria": "MFCC feature extraction is implemented in `src/data_loader.py`.",
            "category": "Data preprocessing and postprocessing",
            "satisfied": null
        },
        {
            "requirement_id": 3,
            "prerequisites": [],
            "criteria": "The \"CNN-LSTM\" model is implemented in 'src/model.py'.",
            "category": "Machine Learning Method",
            "satisfied": null
        },
        {
            "requirement_id": 4,
            "prerequisites": [
                2,
                3
            ],
            "criteria": "Recognition accuracy is saved in `results/metrics/recognition_accuracy.txt`.",
            "category": "Performance Metrics",
            "satisfied": null
        },
        {
            "requirement_id": 5,
            "prerequisites": [
                2,
                3,
                4
            ],
            "criteria": "The confusion matrix is generated and saved as `results/figures/confusion_matrix.png`.",
            "category": "Visualization",
            "satisfied": null
        },
        {
            "requirement_id": 6,
            "prerequisites": [
                2,
                3
            ],
            "criteria": "A local API is created using \"Flask\" to allow users to upload audio files and receive emotion recognition results. The implementation should be included in `src/hci.py`.",
            "category": "Human Computer Interaction",
            "satisfied": null
        }
    ],
    "preferences": [
        {
            "preference_id": 0,
            "criteria": "The audio preprocessing step should be robust, effectively reducing noise while preserving the integrity of the speech signals.",
            "satisfied": null
        },
        {
            "preference_id": 1,
            "criteria": "The local API should be user-friendly, with clear instructions for uploading files and interpreting results.",
            "satisfied": null
        }
    ],
    "is_kaggle_api_needed": true,
    "is_training_needed": true,
    "is_web_navigation_needed": true
}
\end{lstlisting}

\newpage

\section{User experiences of code-generation agentic systems}\label{appendix:user_experiences}


OpenHands~\citep{wang2024opendevin} offers the most refined user experience, leveraging its highly interactive frontend to enable seamless user interaction and task execution. This interface allows users to engage directly with the system, resulting in a smoother and more intuitive workflow, which drives operational efficiency. 

In contrast, MetaGPT~\citep{hongmetagpt} excels in task decomposition through its use of Directed Acyclic Graphs (DAGs), a well-structured and scalable approach aligned with industry best practices in system modularization. This enhances its appeal for users focused on task clarity and modular breakdowns. However, in practical deployments, MetaGPT tends to be less aggressive in file management and preservation, potentially due to its core positioning as a data analysis tool, which does not prioritize persistent state management. Similarly, OpenDevin demonstrates a notable overconfidence in its code generation, frequently skipping the critical step of post-generation code execution, requiring users to intervene manually.

GPT-Pilot~\citep{GPTPilot2023}, praised for its detailed task delegation via over 20 specialized agents, suffers from reduced interactivity due to an overly granular division of responsibilities, resulting in a more fragmented user experience. These qualitative insights, although not fully captured by quantitative metrics, were evident through the DevAI dataset, providing key areas for improvement in user engagement and operational fluidity in future releases of these frameworks.

\newpage

\section{More DevAI dataset samples}\label{appendix:dataset_samples}

\begin{figure}[h!]
    \centering
    \begin{tcolorbox}[
        colframe=metablue,       
        colback=white,         
        coltitle=metablue,       
        title=\textbf{\textcolor{metablue}{Task 13: Style Transfer with Perceptual Loss in PyTorch}},
        fonttitle=\bfseries,   
        boxrule=0.5mm,         
        width=1.1\textwidth,      
        halign=left,         
        colbacktitle=metablue!10 
    ]
    
    \vspace{2pt}
    
    \begin{tcolorbox}[colframe=metablue, colback=metablue!3, sharp corners, boxrule=0.5mm, boxsep=1pt] 
        \begin{center}
            \textbf{\textcolor{metablue}{Query}}
        \end{center}
        \justifying  
        \emph{Please create a PyTorch Perceptual Loss project for image style transfer (refer to this paper: \href{https://arxiv.org/pdf/1603.08155}{Perceptual Losses for Real-Time Style Transfer}). You can build the Perceptual Loss Network using VGG16 in \texttt{src/model.py}. The project should combine content and style images, allow smooth adjustment of style intensity by tuning the weights of style loss and content loss, and save the stylized images in \texttt{results/figures/}. Additionally, log the processing time to \texttt{results/processing\_time.txt}, and save the intermediate results of the style transfer process to \texttt{results/figures/intermediate\_results.png}. For testing, input a famous content image (Mona Lisa) from \href{https://upload.wikimedia.org/wikipedia/commons/thumb/e/ec/Mona_Lisa,_by_Leonardo_da_Vinci,_from_C2RMF_retouched.jpg/768px-Mona_Lisa,_by_Leonardo_da_Vinci,_from_C2RMF_retouched.jpg}{this link} and a famous style image (The Starry Night) from \href{https://ia904705.us.archive.org/34/items/the-starry-night-vincent-van-gogh/The\%20Starry\%20Night\%20-\%20Vincent\%20van\%20Gogh/\%21PREVIEW\%21\%20-\%20The\%20Starry\%20Night\%20-\%20Vincent\%20van\%20Gogh.jpg}{this link}, and generate a style-transferred image. Save the content, style, and style-transferred images to \texttt{data/content.jpg}, \texttt{data/style.jpg}, and \texttt{results/figures/}, respectively. The project should efficiently handle high-resolution images without excessive processing time.}
    \end{tcolorbox}

    \vspace{2pt}
    
    \textbf{\textcolor{metablue}{Requirements}} 
    \begin{itemize}[left=0pt, itemsep=2pt, label={\scriptsize\textcolor{metablue}{$\blacksquare$}}]
        \item \textbf{\textcolor{metablue}{R0}} \\
        \textbf{Criteria:} \emph{A famous content image is inputted for testing, downloaded from \href{https://upload.wikimedia.org/wikipedia/commons/thumb/e/ec/Mona_Lisa,_by_Leonardo_da_Vinci,_from_C2RMF_retouched.jpg/768px-Mona_Lisa,_by_Leonardo_da_Vinci,_from_C2RMF_retouched.jpg}{this link} and saved to \texttt{data/content.jpg}.}\\
        \textbf{Dependencies} \dependencyarrow\;\{\textbf{\textcolor{metablue}{}}\}

        \item \textbf{\textcolor{metablue}{R1}} \\
        \textbf{Criteria:} \emph{A famous style image is inputted for testing, downloaded from \href{https://ia904705.us.archive.org/34/items/the-starry-night-vincent-van-gogh/The\%20Starry\%20Night\%20-\%20Vincent\%20van\%20Gogh/\%21PREVIEW\%21\%20-\%20The\%20Starry\%20Night\%20-\%20Vincent\%20van\%20Gogh.jpg}{this link} and saved in \texttt{data/style.jpg}.} \\
        \textbf{Dependencies} \dependencyarrow\;\{\textbf{\textcolor{metablue}{}}\}

        \item \textbf{\textcolor{metablue}{R2}} \\
        \textbf{Criteria:} \emph{The Perceptual Loss model is implemented in PyTorch and loaded in \texttt{src/model.py}.}\\
        \textbf{Dependencies} \dependencyarrow\;\{\textbf{\textcolor{metablue}{}}\} 

        \item \textbf{\textcolor{metablue}{R3}} \\
        \textbf{Criteria:} \emph{Stylized images are saved to the specified folder \texttt{results/figures/}.}\\
        \textbf{Dependencies} \dependencyarrow\;\{\textbf{\textcolor{metablue}{R0, R1, R2}}\}

        \item \textbf{\textcolor{metablue}{R4}} \\
        \textbf{Criteria:} \emph{Style intensity is adjusted by tuning the weights of style loss and content loss in \texttt{src/model.py}.}\\
        \textbf{Dependencies} \dependencyarrow\;\{\textbf{\textcolor{metablue}{R0, R1, R2}}\}

        \item \textbf{\textcolor{metablue}{R5}} \\
        \textbf{Criteria:} \emph{Processing time is recorded and saved as
        \\\texttt{results/processing\_time.txt}.}\\
        \textbf{Dependencies} \dependencyarrow\;\{\textbf{\textcolor{metablue}{R0, R1, R2, R3, R4}}\}

        \item \textbf{\textcolor{metablue}{R6}} \\
        \textbf{Criteria:} \emph{Intermediate results of style transfer are saved as \texttt{results/figures/intermediate\_results.png}.}\\
        \textbf{Dependencies} \dependencyarrow\;\{\textbf{\textcolor{metablue}{R0, R1, R2, R3, R4}}\}
    \end{itemize}

    \vspace{2pt}

    \textbf{\textcolor{metablue}{Preferences (Optional)}} 
    \begin{itemize}[left=0pt, itemsep=3pt, label={\scriptsize\textcolor{metablue}{$\blacksquare$}}]
        \item \textbf{\textcolor{metablue}{P0}} \\
        \textbf{Criteria:} \emph{The style transfer process should allow for smooth adjustment of style intensity, making the stylized image visually appealing.}
        
        \item \textbf{\textcolor{metablue}{P1}} \\
        \textbf{Criteria:} \emph{The project should handle high-resolution images efficiently without excessive processing time.}
    \end{itemize}


    \end{tcolorbox}
\caption{\textbf{An Example Task in DevAI}: Task 13.}
    \label{fig:task_13}
\end{figure}

\newpage
\begin{figure}[h!]
    \centering
    \begin{tcolorbox}[
        colframe=metablue,       
        colback=white,         
        coltitle=metablue,       
        title=\textbf{\textcolor{metablue}{Task 19: Time Series Forecasting with Seq2Seq LSTM on Rossmann Store Sales}},
        fonttitle=\bfseries,   
        boxrule=0.5mm,         
        width=1.1\textwidth,      
        halign=center,         
        colbacktitle=metablue!10 
    ]
    
    \vspace{1pt}
    
    \begin{tcolorbox}[colframe=metablue, colback=metablue!3, sharp corners, boxrule=0.5mm, boxsep=1pt] 
        \begin{center}
            \textbf{\textcolor{metablue}{Query}}
        \end{center}
        \justifying  
        \emph{Develop a sales forecasting system using a sequence-to-sequence model based on LSTM with the Rossmann Store Sales dataset, downloading it from Kaggle \href{https://www.kaggle.com/c/rossmann-store-sales/data}{here} and loading it in \texttt{src/data\_loader.py}. Split the data into training and testing sets and save them in \texttt{src/data\_loader.py}. Apply a sequence-to-sequence model based on \texttt{LSTM} and save the trained model under the \texttt{models/saved\_models/} directory. Save the forecast results as \texttt{results/figures/forecast\_results.png}. Save a comparison plot between the predicted and actual values to \texttt{results/figures/comparison\_plot.png}. Generate an HTML report that includes the prediction results and comparison plots, with some interactive elements for exploring different forecast horizons, and save it as \texttt{results/report.html}. Ensure the model is tuned to capture seasonal trends in the sales data.}
    \end{tcolorbox}

    \vspace{2pt}
    
    \textbf{\textcolor{metablue}{Requirements}} 
    \begin{itemize}[left=0pt, itemsep=3pt, label={\scriptsize\textcolor{metablue}{$\blacksquare$}}]
        \item \textbf{\textcolor{metablue}{R0}} \\
        \textbf{Criteria:} \emph{The Rossmann Store Sales dataset is used, potentially downloaded from  \href{https://www.kaggle.com/c/rossmann-store-sales/data}{(this link)} and loaded in \texttt{src/data\_loader.py}.}\\
        \textbf{Dependencies} \dependencyarrow\;\{\textbf{\textcolor{metablue}{}}\}

        \item \textbf{\textcolor{metablue}{R1}} \\
        \textbf{Criteria:} \emph{The data is split into training and testing sets and implemented in \texttt{src/data\_loader.py}.}\\
        \textbf{Dependencies} \dependencyarrow\;\{\textbf{\textcolor{metablue}{R0}}\}

        \item \textbf{\textcolor{metablue}{R2}} \\
        \textbf{Criteria:} \emph{A sequence-to-sequence model based on LSTM is used. The trained model should be saved under \texttt{models/saved\_models/}.}\\
        \textbf{Dependencies} \dependencyarrow\;\{\textbf{\textcolor{metablue}{R1}}\}

        \item \textbf{\textcolor{metablue}{R3}} \\
        \textbf{Criteria:} \emph{The forecast results are plotted and saved as \texttt{results/figures/forecast\_results.png}.}\\
        \textbf{Dependencies} \dependencyarrow\;\{\textbf{\textcolor{metablue}{R1, R2}}\}

        \item \textbf{\textcolor{metablue}{R4}} \\
        \textbf{Criteria:} \emph{A comparison plot of predicted vs. actual values is saved as \texttt{results/figures/comparison\_plot.png}.}\\
        \textbf{Dependencies} \dependencyarrow\;\{\textbf{\textcolor{metablue}{R1, R2, R3}}\}

        \item \textbf{\textcolor{metablue}{R5}} \\
        \textbf{Criteria:} \emph{An HTML report containing forecast results and comparison plots is generated and saved as \texttt{results/report.html}.}\\
        \textbf{Dependencies} \dependencyarrow\;\{\textbf{\textcolor{metablue}{R1, R2, R3, R4}}\}

        \item \textbf{\textcolor{metablue}{R6}} \\
        \textbf{Criteria:} \emph{The HTML report should include interactive elements that allow users to explore different forecast horizons.}\\
        \textbf{Dependencies} \dependencyarrow\;\{\textbf{\textcolor{metablue}{R5}}\}
    \end{itemize}

    \vspace{2pt}

    \textbf{\textcolor{metablue}{Preferences (Optional)}} 
    \begin{itemize}[left=0pt, itemsep=3pt, label={\scriptsize\textcolor{metablue}{$\blacksquare$}}]
        \item \textbf{\textcolor{metablue}{P0}} \\
        \textbf{Criteria:} \emph{The model should be tuned to capture seasonal trends in the sales data for more accurate forecasting.}
    \end{itemize}


    \end{tcolorbox}
\caption{\textbf{An Example Task in DevAI}: Task 19.}
    \label{fig:task_19}
\end{figure}

\newpage

\section{Human Evaluation Procedure}\label{appendix:human_eval}
We recruited three AI experts from the authors to perform human evaluation on the output of agentic code generation systems. There we present the evaluation details.

\paragraph{First round} 
For the first round of evaluations, our three evaluators reported spending $16.5$, $19.5$, and $22.0$ hours, respectively.
To capture the bias that a human evaluator will have, the instructions given to our experts were minimal, with them only receiving a scorecard to complete for each agentic system and each task.
Results that all evaluators agree on are considered trustworthy. The assumption here is that it is unlikely that all three evaluators make a mistake or have an effective bias in the same judgment. 
The self-reported post-hoc evaluation criteria are shown in Figure~\ref{fig:evaluation_criteria}.

\paragraph{Second round} In the second round, the evaluators present and discuss their reasons for disagreeing with judges. In doing so, human errors are likely corrected by their peers. Discussion among evaluators also helps reduce human bias by examining each other's thought processes thoroughly. Furthermore, the consensed results are considered trustworthy given the assumption that it is unlikely that all three evaluators are convinced by the same mistake or the same cognitive bias. The three evaluators took 9.5 hours together for this second round of evaluation.

\newpage
\begin{figure}[h!]
\vspace{-1cm}
\small
    \centering
    \begin{tcolorbox}[
        colframe=metablue,       
        colback=white,         
        coltitle=metablue,       
        title=\textbf{\textcolor{metablue}{Self-reported Post-hoc Evaluation Criteria after round one}},
        fonttitle=\bfseries,   
        boxrule=0.5mm,         
        width=\textwidth,      
        halign=center,         
        colbacktitle=metablue!10 
    ]
    

    \begin{tcolorbox}[colframe=metablue, colback=metablue!3, sharp corners, boxrule=0mm, boxsep=1pt] 
        \begin{center}
            \textbf{\textcolor{metablue}{Evaluator $231a$}}
        \end{center}
        \justifying  
\texttt{EXECUTED SUCCESSFULLY: Yes [ ] / No [ ]}
\\ 
1. Must be checked based on the overall completeness of the task, based on looking at the code, the artifacts, and the trajectory. \\
2. The training has finished, the model snapshot and the metrics breakdown have been saved, and at least one artifact of required analytics has been produced - mark as successful. If some analytic artifacts are missing but not all, mark as successful.\\
3. No need to run the code.\\
4. If training was finished but on fake data, mark as successful.\\
\texttt{Requirements:}\\
Marking a requirement as satisfied must be made for this specific requirement disregarding the dependency list.
If a file (code, image, snapshot) is there but is empty or without any meaningful content - mark as No.\\
1. Code:\\
The functionality must be in a file with the requested path.\\
The real data is replaced by simple synthetic - not satisfied.\\
2. Visualization/Reports:\\
The contents must be there and make sense even if not perfect from the ML/DS point of view.\\
3. Snapshots:\\
If a binary snapshot is not empty, mark as Yes.
\end{tcolorbox}

\newpage
     \begin{tcolorbox}[colframe=metablue, colback=metablue!3, sharp corners, boxrule=0mm, boxsep=1pt] 
        \begin{center}
            \textbf{\textcolor{metablue}{Evaluator $38bb$}}
        \end{center}
        \justifying  
\texttt{EXECUTED SUCCESSFULLY: Yes [ ] / No [ ]}\\
An output is marked yes if none of the following is satisfied.\\
1. The time spent is close to the time limit.\\
2. The last environmental message includes an error.\\
3. The last thought indicates that the task is completed. \\
4. The last step of the trajectory is incomplete.\\
\texttt{Requirements:}\\
If a required is unsatisfied only because of an unsatisification of a previously marked unsatisfied requirment, then judge it based on the assumption that a minimum implementation satisfies the previous requirment exists.\\
1. Code:\\
Mark yes if the code executes and does the required function. If no entrypoint is given, the evaluator will set an entrypoint. If the code is not executable due to previous unsatisfied requirements, then it is judged based on eye-checking.\\
2. Visualization/Reports:\\
Mark yes if the visualization or report exists in a right path and the content aligns the requirement. \\
3. Snapshots:\\
Mark yes if the snapshot exists in a right path and is not empty.
    \end{tcolorbox}
     \begin{tcolorbox}[colframe=metablue, colback=metablue!3, sharp corners, boxrule=0mm, boxsep=1pt] 
        \begin{center}
            \textbf{\textcolor{metablue}{Evaluator $cn9o$}}
        \end{center}
        \justifying  

I evaluated everything based on whether the requirement was satisfied verbatim, using my own professional judgement when there was ambiguity. If there was a data folder and a look at the trajectory indicated it was real, I allowed it to be loaded directly. If there were results and no clear source code making them, I considered the task incomplete. I didn't consider a dummy data path to be correct (GPT-Pilot used a lot of these). To determine if something was executed correctly, I skimmed the end of the logs (e.g., trajectory) for any obvious signs of an error. If there wasn't any, I said it was correctly executed. I ignored prerequisites while evaluating and did not execute anything, instead just eyeballing the code for correctness. I was lenient in what I considered to be sufficient in terms of the more vague requirements (e.g., if the preprocessing had to include scaling and rotation, just those two would be sufficient to consider it done).

    \end{tcolorbox}

    \end{tcolorbox}
    \vspace{-7pt}
\caption{Each evaluator was given a full description of each task and the associated requirements and preferences in markdown format. They were then asked, for each workspace and trajectory generated by each of the agents on each of the task, whether (1) the agent successfully finished its execution cycle and (2) which of the requirements were satisfied. After the evaluation was complete, the evaluators were asked to self-report the nuances of their evaluation.}
    \label{fig:evaluation_criteria}
    \vspace{-10pt}
\end{figure}

\newpage

\section{Suggest Constraints}\label{appendix:constraint}

\noindent Below is a sample of constraints in JSON format that describes task-specific guidelines:

\begin{lstlisting}[language=json]
{
   "generic": "This is a task that requires you to write, execute, and save source code. You have a hard time limit of 30 minutes to produce your programmatic solution to the given task. This time limit includes execution time. The quality of your solution will be judged based on what you left in the working folder by the time 30 minutes expire. Additionally, the hardware you are running on is unknown, and the presence of a GPU is not guaranteed.",
   "is_training_needed": "Keep the time limit in mind when setting hyperparameters for training.",
   "is_kaggle_api_needed": "You can use the Kaggle API credentials stored in `kaggle.json` in your current working directory."
}
\end{lstlisting}

To address automation and security concerns, we have written code to place the \texttt{kaggle.json} file into the current workspace for each baseline during each run. 
However, dataset users are free to modify the solution to enable the Kaggle API for their specific developer agents.

\newpage

\section{Collected Trajectories}\label{appendix:trajectory}

\subsection{Schema}
\noindent Below is the required JSON format for a trajectory in gray-box settings (where the trajectories can serve as input for LLM-as-a-Judge and Agent-as-a-Judge).

\vspace{10pt}

\begin{lstlisting}[language=json]
{
  "type": "array",
  "items": {
    "type": "object",
    "properties": {
      "step": {
        "type": "integer",
        "description": "The step number in the trajectory, 0-based."
      },
      "user_message": {
        "type": ["string", "null"],
        "description": "The message from the external user to the agent. If null, no message was sent."
      },
      "agent": {
        "type": "object",
        "properties": {
          "thought": {
            "type": "string",
            "description": "The agent's thought at this step."
          },
          "action": {
            "type": ["string", "null"],
            "description": "The agent's action sent to the environment. If null, the agent did not take any action, for example, when the agent has finished the task."
          },
          "agent_name": {
            "type": "string",
            "description": "The name of the agent that made the action."
          }
        },
        "required": ["thought", "action"],
        "description": "Everything related to the agent at this step."
      },
      "environment": {
        "type": ["string", "null"],
        "description": "The environment's (shell, python interpreter) response to the action submitted by the agent. If null, the environment was not involved in this step."
      },
      "step_usage": {
        "type": "object",
        "properties": {
          "input_tokens": {
            "type": "integer",
            "description": "The number of input tokens passed as LLM context."
          },
          "output_tokens": {
            "type": "integer",
            "description": "The number of tokens produced by the LLM."
          },
          "model": {
            "type": "string",
            "description": "The name of the LLM model used."
          },
          "cost": {
            "type": "number",
            "description": "The cost of the LLM inference, in USD."
          },
          "llm_inference_time": {
            "type": "number",
            "description": "The time taken by the LLM to generate the output tokens, in seconds."
          },
          "step_execution_time": {
            "type": "number",
            "description": "The time taken to make an entire step including LLM inference and environment execution, in seconds."
          }
        },
        "required": [
          "input_tokens",
          "output_tokens",
          "model",
          "cost",
          "llm_inference_time",
          "step_execution_time"
        ]
      },
      "accumulated_usage": {
        "type": "object",
        "properties": {
          "accumulated_cost": {
            "type": "number",
            "description": "The total cost of the trajectory up to this step, in USD."
          },
          "accumulated_time": {
            "type": "number",
            "description": "The total time taken by the agent to complete the trajectory up to this step, in seconds."
          }
        },
        "required": [
          "accumulated_cost",
          "accumulated_time"
        ]
      }
    },
    "required": ["step", "user_message", "agent", "environment", "step_usage", "accumulated_usage"]
  }
}
\end{lstlisting}

\newpage

\subsection{A Sample of Trajectory}

This is sample trajectories (we select 4 steps of it) in OpenHands~\citep{wang2024opendevin}. All steps are recorded to enable replay of the trajectory.

\vspace{10pt}

\begin{lstlisting}[language=json]
{
        "step": 0,
        "user_message": "Help me develop an object detection system using the YOLOv3 model and the COCO dataset. Download the dataset and preprocess the images by resizing and normalization in `src/data_loader.py`. Implement the YOLOv3 model and use Non-Maximum Suppression (NMS) to refine the results in `src/model.py`. Save the detected objects to `results/figures/`, and create an interactive Streamlit web page in `src/app.py` to display the detection results. Finally, evaluate the model's performance, including metrics such as mAP and inference time, and save the evaluation results to `results/metrics/model_performance.txt`. The system should properly manage the launch and termination of the Streamlit application to prevent unnecessary resource usage.\n\nThis is a task that requires you to write, execute, and save source code. You have a hard time limit of 30 minutes to produce your programmatic solution to the given task. This time limit includes execution time. The quality of your solution will be judged based on what you left in the working folder by the time 30 minutes expire. Additionally, the hardware you are running on is unknown, and the presence of a GPU is not guaranteed.\n\nKeep the time limit in mind when setting hyperparameters for training.\n",
        "agent": {
            "thought": "Let's break down the task into manageable steps and start implementing them one by one.\n\n### Steps:\n1. **Set up the directory structure**:\n   - Create necessary directories: `src/` and `results/figures/`.\n2. **Download and preprocess the COCO dataset**:\n   - Implement `src/data_loader.py` to download, resize, and normalize images.\n3. **Implement the YOLOv3 model**:\n   - Implement `src/model.py` to define the YOLOv3 model and apply Non-Maximum Suppression (NMS).\n4. **Create an interactive Streamlit web page**:\n   - Implement `src/app.py` to display detection results.\n5. **Evaluate the model's performance**:\n   - Save evaluation metrics to `results/metrics/model_performance.txt`.\n\nLet's start by setting up the directory structure and creating the necessary files.\n\n### Step 1: Set up the directory structure",
            "action": "Running command: mkdir -p src results/figures results/metrics\ntouch src/data_loader.py src/model.py src/app.py"
        },
        "environment": "\r\n[Python Interpreter: /openhands/poetry/openhands-5O4_aCHf-py3.11/bin/python]\nopenhands@b34322eae8da:/workspace $ touch src/data_loader.py src/model.py src/app.py\r\n\r\n[Python Interpreter: /openhands/poetry/openhands-5O4_aCHf-py3.11/bin/python]\nopenhands@b34322eae8da:/workspace $ ",
        "step_usage": {
            "input_tokens": 4331,
            "output_tokens": 220,
            "llm_inference_time": 4.445789337158203,
            "model": "gpt-4o-2024-05-13",
            "cost": 0.024955,
            "step_execution_time": 16.24921226501465
        },
        "accumulated_usage": {
            "accumulated_cost": 0.024955,
            "accumulated_time": 16.24922204017639
        }
    },
    {
        "step": 1,
        "user_message": null,
        "agent": {
            "thought": "Great! The directory structure and necessary files are created. Let's proceed with implementing the data loader to download and preprocess the COCO dataset.\n\n### Step 2: Implement `src/data_loader.py`\nWe will:\n1. Download the COCO dataset.\n2. Resize and normalize the images.\n\nLet's start by writing the code for `src/data_loader.py`.",
            "action": "Running Python code interactively: create_file('src/data_loader.py')"
        },
        "environment": "FileExistsError                           Traceback (most recent call last)\nCell In[1], line 1\n----> 1 create_file('src/data_loader.py')\n\nFile /openhands/code/openhands/runtime/plugins/agent_skills/file_ops/file_ops.py:257, in create_file(filename)\n    251 \"\"\"Creates and opens a new file with the given name.\n    252 \n    253 Args:\n    254     filename: str: The name of the file to create.\n    255 \"\"\"\n    256 if os.path.exists(filename):\n--> 257     raise FileExistsError(f\"File '{filename}' already exists.\")\n    259 with open(filename, 'w') as file:\n    260     file.write('\\n')\n\nFileExistsError: File 'src/data_loader.py' already exists.\n[Jupyter current working directory: /workspace]\n[Jupyter Python interpreter: /openhands/poetry/openhands-5O4_aCHf-py3.11/bin/python]",
        "step_usage": {
            "input_tokens": 4675,
            "output_tokens": 84,
            "llm_inference_time": 2.136143207550049,
            "model": "gpt-4o-2024-05-13",
            "cost": 0.024635000000000004,
            "step_execution_time": 3.345384359359741
        },
        "accumulated_usage": {
            "accumulated_cost": 0.04959000000000001,
            "accumulated_time": 19.594608068466187
        }
    },
    {
        "step": 2,
        "user_message": null,
        "agent": {
            "thought": "The file `src/data_loader.py` already exists. Let's open it and start writing the code to download and preprocess the COCO dataset.\n\n### Step 2: Implement `src/data_loader.py`",
            "action": "Running Python code interactively: open_file('src/data_loader.py')"
        },
        "environment": "[File: /workspace/src/data_loader.py (1 lines total)]\n(this is the beginning of the file)\n1|\n(this is the end of the file)\n[Jupyter current working directory: /workspace]\n[Jupyter Python interpreter: /openhands/poetry/openhands-5O4_aCHf-py3.11/bin/python]",
        "step_usage": {
            "input_tokens": 4982,
            "output_tokens": 53,
            "llm_inference_time": 2.209756851196289,
            "model": "gpt-4o-2024-05-13",
            "cost": 0.025705000000000002,
            "step_execution_time": 2.318861961364746
        },
        "accumulated_usage": {
            "accumulated_cost": 0.07529500000000001,
            "accumulated_time": 21.913471937179565
        }
    },
    ...
    {
        "step": 20,
        "user_message": null,
        "agent": {
            "thought": "It seems the indentation issue persists. Let's correct the indentation manually.\n\n### Corrected Code for `src/evaluate.py`\n```python\nimport time\nimport numpy as np\nfrom model import yolo_v3_model, non_max_suppression\nfrom data_loader import preprocess_image, download_coco_dataset, extract_coco_dataset\nimport os\n\ndef evaluate_model():\n    model = yolo_v3_model()\n    download_coco_dataset()\n    extract_coco_dataset()\n\n    image_dir = \"data/val2017\"\n    image_files = [os.path.join(image_dir, f) for f in os.listdir(image_dir) if f.endswith('.jpg')]\n\n    total_inference_time = 0\n    num_images = len(image_files)\n    all_boxes = []\n    all_scores = []\n    all_classes = []\n\n    for image_file in image_files:\n        image = preprocess_image(image_file)\n        image = np.expand_dims(image, axis=0)\n\n        start_time = time.time()\n        predictions = model.predict(image)\n        end_time = time.time()\n\n        inference_time = end_time - start_time\n        total_inference_time += inference_time\n\n        boxes, scores, classes = non_max_suppression(predictions[0])\n        all_boxes.append(boxes)\n        all_scores.append(scores)\n        all_classes.append(classes)\n\n    mean_inference_time = total_inference_time / num_images\n\n    # Placeholder for mAP calculation\n    mAP = 0.0  # This should be replaced with actual mAP calculation\n\n    with open(\"results/metrics/model_performance.txt\", \"w\") as f:\n        f.write(f\"Mean Inference Time: {mean_inference_time:.4f} seconds\\n\")\n        f.write(f\"mAP: {mAP:.4f}\\n\")\n\nif __name__ == \"__main__\":\n    evaluate_model()\n```\n\nLet's update the file with the corrected code.",
            "action": "Running Python code interactively: EDITED_CODE = \"\"\"import time\nimport numpy as np\nfrom model import yolo_v3_model, non_max_suppression\nfrom data_loader import preprocess_image, download_coco_dataset, extract_coco_dataset\nimport os\n\ndef evaluate_model():\n    model = yolo_v3_model()\n    download_coco_dataset()\n    extract_coco_dataset()\n\n    image_dir = \"data/val2017\"\n    image_files = [os.path.join(image_dir, f) for f in os.listdir(image_dir) if f.endswith('.jpg')]\n\n    total_inference_time = 0\n    num_images = len(image_files)\n    all_boxes = []\n    all_scores = []\n    all_classes = []\n\n    for image_file in image_files:\n        image = preprocess_image(image_file)\n        image = np.expand_dims(image, axis=0)\n\n        start_time = time.time()\n        predictions = model.predict(image)\n        end_time = time.time()\n\n        inference_time = end_time - start_time\n        total_inference_time += inference_time\n\n        boxes, scores, classes = non_max_suppression(predictions[0])\n        all_boxes.append(boxes)\n        all_scores.append(scores)\n        all_classes.append(classes)\n\n    mean_inference_time = total_inference_time / num_images\n\n    # Placeholder for mAP calculation\n    mAP = 0.0  # This should be replaced with actual mAP calculation\n\n    with open(\"results/metrics/model_performance.txt\", \"w\") as f:\n        f.write(f\"Mean Inference Time: {mean_inference_time:.4f} seconds\\\\n\")\n        f.write(f\"mAP: {mAP:.4f}\\\\n\")\n\nif __name__ == \"__main__\":\n    evaluate_model()\n\"\"\"\n\nedit_file_by_replace('src/evaluate.py', 'with open(\"results/metrics/model_performance.txt\", \"w\") as f:', '    with open(\"results/metrics/model_performance.txt\", \"w\") as f:')"
        },
        "environment": null,
        "step_usage": {
            "input_tokens": 20132,
            "output_tokens": 805,
            "llm_inference_time": 15.1565523147583,
            "model": "gpt-4o-2024-05-13",
            "cost": 0.11273500000000002,
            "step_execution_time": 15.196227073669434
        },
        "accumulated_usage": {
            "accumulated_cost": 1.2919750000000003,
            "accumulated_time": 176.99772310256958
        }
    }
\end{lstlisting}

\newpage
\section{Ablations of Agent-as-a-Judge}\label{appendix:ablation_aaaj}

\subsection{Component Abalations}

\begin{table}[ht]
  \centering
  \caption{
  \textbf{Component Ablation Studies for Agent-as-a-Judge.} We analyze the impact of sequentially adding various components (\texttt{ask}, \texttt{graph}, \texttt{read}, \texttt{locate}, \texttt{retrieve}, \texttt{planning}, \texttt{memory}, and \texttt{search}) on the performance of Agent-as-a-Judge for judging OpenHands.  }
  \label{tab:agent_judge_ablation_2}
  \renewcommand\tabcolsep{4pt} 
  \renewcommand\arraystretch{1.5} 
  \resizebox{1\textwidth}{!}{%
    \begin{tabular}{lccccccccc} 
      \Xhline{1.4pt}
      \rowcolor{metablue!20} 
      \textbf{Metric} & + \texttt{ask} & + \texttt{graph} & + \texttt{read} & + \texttt{locate} & + \texttt{search} & + \texttt{retrieve} & + \texttt{planning} & + \texttt{memory} \\
      \Xhline{1.5pt}
      \rowcolor{gray!10}  \multicolumn{9}{c}{\textbf{Agent-as-a-Judge Performance}} \\
      \cline{1-8}  
      \Xhline{1.5pt}
      \textbf{Alignment Rate} & 65.03\% & 75.95\% & 82.24\% & 90.44\% & 86.06\% & 90.16\% & 88.52\% & 87.97\% \\
      \Xhline{1.5pt}
    \end{tabular}
  }
\end{table}

\noindent \textbf{Analysis} 
We designed 8 modular components for the Agent-as-a-Judge system. In the \Cref{tab:agent_judge_ablation_2}, components are added progressively from left to right. If the addition of a component led to a significant performance drop, we removed it from further iterations. Our experiments showed that adding the components \texttt{ask}, \texttt{graph}, \texttt{read}, and \texttt{locate} resulted in significant performance gains. However, when the \texttt{search} component was introduced, there was a noticeable decline in performance.

We hypothesize that the performance drop from \texttt{search} is due to its role in retrieving relevant code snippets (top-$3$) using BM25. The retrieval accuracy of BM25~\citep{robertson2009probabilistic} might not have been high enough, potentially introducing noise. Moreover, as noted in \Cref{tab:ai_dev_comparison_1}, the DevAI tasks in our experiments did not generate a large amount of code. In fact, even when all code was fed into an LLM, the total content typically stayed within the maximum context length. Therefore, in simpler workspaces, \texttt{search} was less critical. However, we believe this component will become more important as the complexity of the workspace increases, making it more valuable in larger and more complex environments.

We also observed that the introduction of the \texttt{planning} mechanism did not bring a noticeable improvement in performance. This may be related to the nature of the Judge - it needs clean factual information. When \texttt{planning} is unstable, the evidence collected from different actions can become inconsistent, leading to a decline in performance. Finally, we experimented with a \texttt{memory} mechanism. Initially, we hypothesized that since DevAI tasks often involve interconnected requirements, \texttt{memory} could help track whether requirements were met. However, in practice, we saw no improvement. We suspect that the interconnected nature of the requirements may have caused biases: specifically, once a prior requirement was fulfilled, it might have overly influenced positive judgments on subsequent requirements, even if they were not fully met.

\subsection{Search Algorithms in Search Module}

We initially hypothesized that the performance drop was due to the low precision of the \texttt{search} component, particularly with BM2.5. To explore this, we replaced BM2.5 with Sentence-BERT~\citep{reimers2019sentence} as a more advanced alternative and tested Fuzzy Search~\citep{levenshtein1966binary} as a less precise option. However, neither improved the performance of the Agent-as-a-Judge.

\begin{wraptable}{r}{0.35\textwidth}
  \centering
  \vspace{-20pt}
  \captionof{table}{
  \small Comparisons on \texttt{Search} module with different engines.}
  \label{tab:search_comparison}
  \renewcommand\tabcolsep{2.1pt}
  \renewcommand\arraystretch{1.3} 
 \resizebox{0.35\textwidth}{!}{%
  \begin{tabular}{lccc} 
    \Xhline{1.4pt}
    \rowcolor{metablue!20} 
    \multicolumn{1}{c}{\textbf{Search Method}} & \textbf{Alignment Rate} \\
    \Xhline{1.5pt}
    \rowcolor{gray!10} BM2.5 &  86.06\% \\
    Sentence-BERT &  87.70\% \\
    \rowcolor{gray!10} Fuzzy Search & 85.52\% \\
    \emph{without} Search Module &  90.44\% \\
    \Xhline{1.5pt}
  \end{tabular}
 }
 \vspace{-5pt}
\end{wraptable}

hese results suggest that the performance issue is not due to BM2.5's poor search accuracy. Instead, the workspaces generated in our DevAI tasks are too simple for the \texttt{search} component to have a significant impact. In simpler workspaces, direct retrieval and evaluation are sufficient. Even though Sentence-BERT performed better than the other methods, its alignment rate ($87.70\%$) still falls short of the configuration without the \texttt{search} component ($90.44\%$). As workspace complexity increases, the \texttt{search} component may become more valuable.

\subsection{Search Algorithms in Retrieve Module}
In our experiments, we found that accurately locating relevant information within a trajectory is a challenging task. Although the addition of the \texttt{retrieve} component (gray-box) did not lead to a significant improvement in performance in this specific case, its impact has been notable in other settings, such as in GPT-Pilot. As shown in \Cref{tab:ai_dev_comparison_alignment}, the integration of \texttt{retrieve} in GPT-Pilot brought substantial gains.

\begin{wraptable}{r}{0.4\textwidth}
  \centering
  \vspace{-15pt}
  \captionof{table}{
  \small Ablations on \texttt{retrieve}.}
  \label{tab:retrieve_ablation}
  \renewcommand\tabcolsep{2.1pt}
  \renewcommand\arraystretch{1.3} 
 \resizebox{0.4\textwidth}{!}{%
  \begin{tabular}{lcc} 
    \Xhline{1.4pt}
    \rowcolor{metablue!20} 
    \multicolumn{1}{c}{\textbf{Method}} & \textbf{Alignment Rate} \\
    \Xhline{1.5pt}
    \rowcolor{gray!10} Without \texttt{retrieve} &  83.88\% \\
    With \texttt{retrieve} (gray-box) &  86.61\% \\
    \rowcolor{gray!10} Trajectory Truncate (head) & 86.61\% \\
    Trajectory Truncate (middle) & 85.52\% \\
    \rowcolor{gray!10} Trajectory Truncate (tail) & 82.51\% \\
    Step Truncate (head) & 86.34\% \\
    \rowcolor{gray!10} Step Truncate (middle) &86.61\% \\
    Step Truncate (tail) & 83.88\% \\
    \Xhline{1.5pt}
  \end{tabular}
 }
\end{wraptable}

We conducted an ablation study on GPT-Pilot to optimize the retrieval of useful information at each step. Our experiments revealed that in large trajectories, truncating the final sections of the file often results in losing critical information, as the latter part of the trajectory typically contains dense information about the final development state. Truncating the beginning of the trajectory proved to be the most effective in improving the retrieval efficiency.

For individual steps, truncating the middle section worked best. This is because error messages usually appear early in the output, while the corresponding file paths and specific error locations are found towards the end. By focusing on these retrieval strategies, we can significantly enhance the performance of the \texttt{retrieve} component, particularly in complex scenarios like GPT-Pilot.
\newpage
\section{Prompt Demos of Agent-as-a-Judge}\label{appendix:judge_agent_prompt}

Here, we present some prompts used by the Agent-as-a-Judge system. Each of these prompt demos plays a crucial role in guiding the agent's behavior.

\subsection{System Prompt for Agent-as-a-Judge}

\begin{lstlisting}[language=custompython]
def get_system_prompt(language="English"):

    if language == "English":
        return """
        You are an advanced AI system serving as an impartial judge for intelligent code generation outputs. Your primary role is to rigorously evaluate whether the agent's outputs satisfy the specified requirements by thoroughly analyzing the provided code, data, and other relevant materials.

        You will systematically assess aspects such as datasets, model implementations, training procedures, and any task-specific criteria outlined in the requirements. Your evaluations must be objective, detailed, and based solely on the evidence provided.

        For each requirement, deliver one of the following judgments:

        1. <SATISFIED>: Use this if the agent's output fully meets the requirement. Provide a brief and precise explanation demonstrating how the specific criteria are fulfilled.

        2. <UNSATISFIED>: Use this if the agent's output does not meet the requirement. Provide a concise explanation indicating the deficiencies or omissions.

        Your assessment should reference specific elements such as code snippets, data samples, or output results where appropriate. Ensure that your justifications are clear, precise, and directly related to the criteria.

        Respond with either <SATISFIED> or <UNSATISFIED>, followed by your concise justification.
        """
    else:
        raise NotImplementedError(f"The language '{language}' is not supported.")

\end{lstlisting}

\newpage
\subsection{System Prompt for Locate Module}

\begin{lstlisting}[language=custompython]
def get_locate_system_prompt(language="English"):
    if language == "English":
        return """
        You are an advanced AI system specializing in understanding project structures and determining file locations based on provided criteria.
        Your task is to locate specific files in the workspace based on the user's criteria and workspace information.ution problems with the files mentioned in the criteria.
        """

    else:
        raise NotImplementedError(f"The language '{language}' is not supported.")
\end{lstlisting}

\subsection{System Prompt for Retrieve Module}

\begin{lstlisting}[language=custompython]
def get_retrieve_system_prompt(language="English"):

    if language == "English":
        return """
        You are an advanced AI system specializing in retrieving environmental feedback from project execution trajectories. Your task is to analyze the provided trajectory data and extract information about the most relevant files mentioned in the given criteria.
        
        Focus on the following:
        
        1. Identify the **most recent steps** where the files directly related to the criteria were involved in execution, loading, or saving operations.
        2. Provide environmental feedback for these files, such as any errors, warnings, or issues encountered during their execution or processing.
        3. Highlight whether any problems occurred that might affect the functionality or success of these files in the project.
        
        Your output should be structured as follows:
        
        - **<RELEVANT STEPS>**: List the specific steps involving the relevant files, including any environmental feedback such as error messages, execution results, or other issues encountered. Each step should concisely present the key information needed to assess the files' execution status.
        
        Avoid including details about file contents or existence, as this information is already available. Focus solely on the environmental feedback related to the execution of the most relevant files.
        
        Your goal is to provide clear and concise information that helps determine if there were any execution problems with the files mentioned in the criteria.
        """
    else:
        raise NotImplementedError(f"The language '{language}' is not supported.")

\end{lstlisting}

\newpage
\subsection{Prompt for Ask Module (for requirement check)}
\begin{lstlisting}[language=custompython]
def get_ask_prompt(criteria: str, evidence: str) -> str:

    return f"""
    Provided below is relevant information about the project:
    {evidence}
    
    Kindly perform an evaluation of the following criteria:
    {criteria}
    
    As per the guidelines, respond with either <SATISFIED> or <UNSATISFIED>, followed by a concise justification that references specific elements from the project information, such as code snippets, data samples, or output results.
    """

\end{lstlisting}

\subsection{Prompt for Locate Module}

\begin{lstlisting}[language=custompython]
def get_locate_prompt(criteria: str, evidence: str) -> str:

    return f"""
    Provided below is the structure of the workspace:
    {workspace_info}

    This is the criteria related to the task:
    {criteria}

    Follow the format in the example below and return only the file paths that match the criteria:
    
    Example:
    
    Suppose the criteria is:
    'The database functionality is implemented in `src/db.py`, and the logging system is defined in `src/logging.py`.'

    And the workspace information is:
    /project
    |-- src
    |   |-- db.py
    |   |-- logging.py
    |   |-- utils.py
    |-- tests
        |-- test_db.py
        |-- test_logging.py

    Based on the criteria, the following paths (no more than 5) should be returned, each wrapped in dollar signs (`$`):
    $/project/src/db.py$
    $/project/src/logging.py$
    """
\end{lstlisting}

\newpage
\section{Judge Evidences Collected from Agent-as-a-Judge}\label{appendix:judge_agent_evidence}

\begin{figure}[h]
\centering
\includegraphics[width=0.561\linewidth]{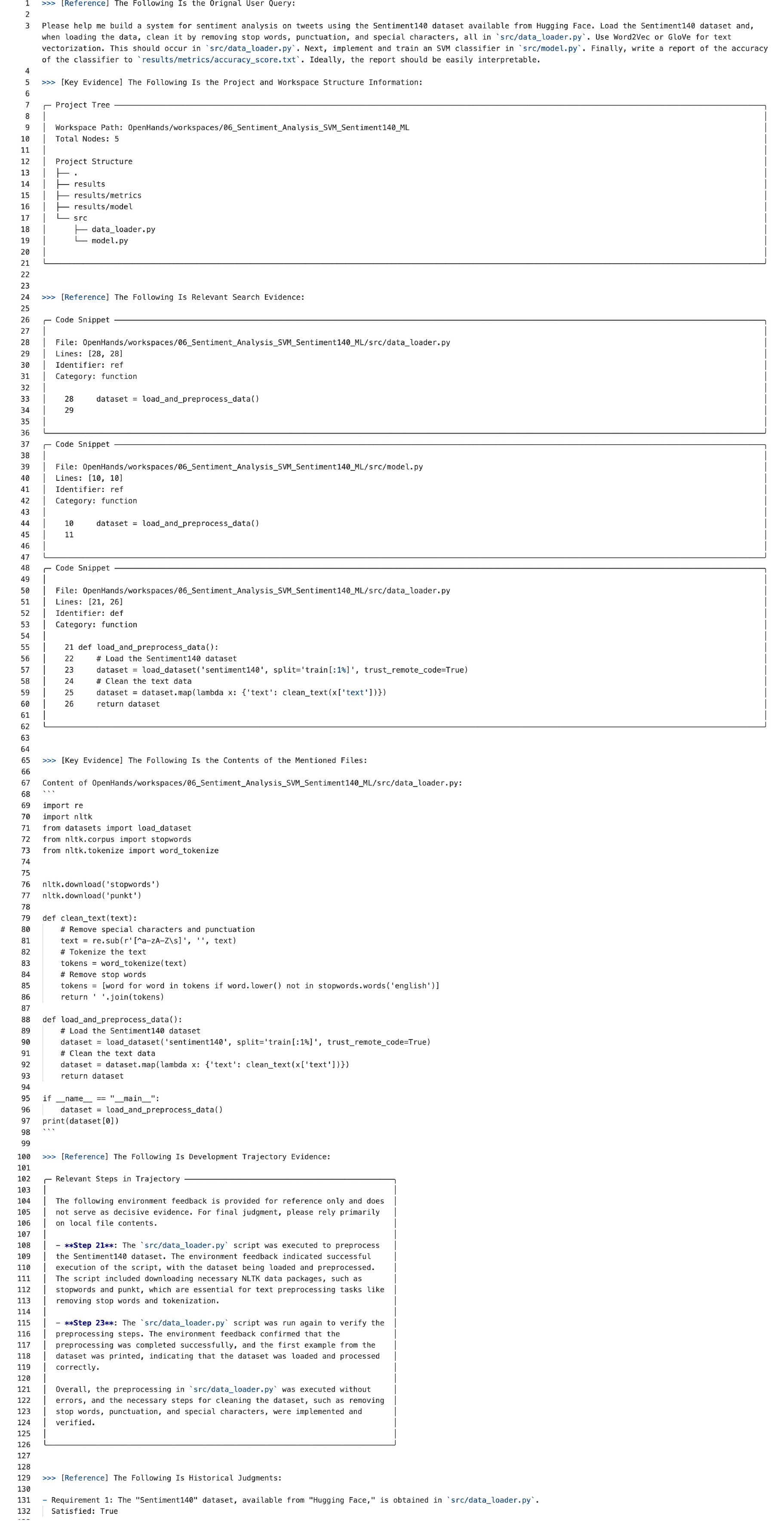}
\label{fig:evidence}
\end{figure}

The Agent-as-a-Judge system dynamically collects evidence throughout the development process of the code-generating agentic system to evaluate whether each requirement is satisfied. Using this auxiliary information, the judge agent decides to focus on relevant data such as code, file structures, and trajectories. This gathered evidence supports the agent's final judgment, ensuring an informed and accurate assessment of the project's compliance with its requirements. A sample of the collected evidence is shown in the following \Cref{fig:evidence}.

\end{document}